\documentclass[arxiv]{melba}

\usepackage{mwe} 

\usepackage{rotating} 
\usepackage{pdflscape}

\usepackage{amsmath,amsfonts}

\usepackage[table]{xcolor}

\usepackage{amssymb} 
\usepackage{pifont}

\usepackage{booktabs}
\usepackage{tabularx}
\usepackage{makecell}
\usepackage[table]{xcolor} 
\usepackage{multirow}
\usepackage{threeparttable}
\usepackage{multirow}

\usepackage{booktabs}
\usepackage{array}
\usepackage{makecell}
\usepackage[table]{xcolor}
\usepackage{multirow}
\usepackage{threeparttable}

\melbaid{YYYY:NNN}  
\doi{10.59275/j.melba.2024-AAAA}
\melbaauthors{Xing Yao, Ahana Gangopadhyay, Hsi-Ming Chang, Ravi Son}  
\email{Ahana.Gangopadhyay@gehealthcare.com}
\volume{3}
\firstpageno{1337}  
\melbayear{YYYY}  
\datesubmitted{yyyy-m1-d1}  
\datepublished{yyyy-m2-d2}  

\ShortHeadings{MELBA Journal Sample Article}{Xing Yao, Ahana Gangopadhyay, Hsi-Ming Chang, Ravi Soni}

\title{Towards Better Ultrasound Video Segmentation Foundation Model: An Empirical study on SAM2 Finetuning from Data Perspective}

\author{
	\firstname Xing \surname Yao\aff{1,2},
	\firstname Ahana \surname Gangopadhyay\aff{1},
        \firstname Hsi-Ming \surname Chang\aff{1},
        \firstname Ravi \surname Soni\aff{1}
}
\affiliations{
	\num 1 \addr GE HealthCare, San Ramon, USA \\
	\num 2 \addr Vanderbilt University, Nashville, USA \\
}

\abstract{
Ultrasound (US) video segmentation remains a challenging problem due to strong inter- and intra-dataset variability, motion artifacts, and limited annotated data. Although foundation models such as Segment Anything Model 2 (SAM2) demonstrate strong zero-shot and prompt-guided segmentation capabilities, their performance deteriorates substantially when transferred to medical imaging domains. Current adaptation studies mainly emphasize architectural modifications, while the influence of data characteristics and training regimes has not been systematically examined. In this study, we present a comprehensive, data-centric investigation of SAM2 adaptation for ultrasound video segmentation. We analyze how training-set size, video duration, and augmentation schemes affect adaptation performance under three paradigms: task-specific fine-tuning, intermediate adaptation, and multi-task joint training, across five SAM2 variants and multiple prompting modes. We further design six ultrasound-specific augmentations, assessing their effect relative to generic strategies. Experiments on three representative ultrasound datasets reveal that data scale and temporal context play a more decisive role than model architecture or initialization. Moreover, joint training offers an efficient compromise between modality alignment and task specialization. This work aims to provide empirical insights for developing efficient, data-aware adaptation pipelines for SAM2 in ultrasound video analysis.
}


\begin{document}

\twocolumn[\maketitle]

\section{Introduction}
\label{sec:intro}
\begin{figure*}[t]
    \includegraphics[width=1.0\textwidth]{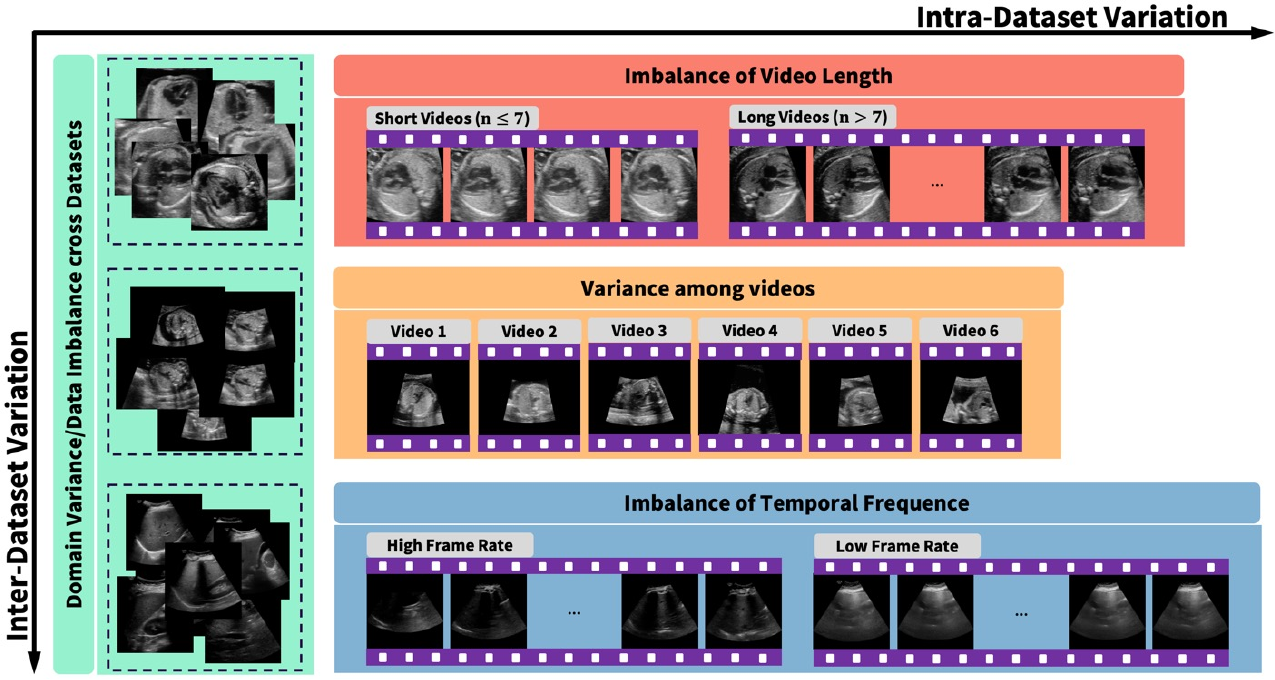}
    \caption{Illustration of data variation in medical ultrasound video datasets. The figure highlights both inter-dataset and intra-dataset variability, including domain shifts across datasets, class imbalance, and differences in video length, temporal resolution, and content diversity within a single dataset. These factors can significantly influence fine-tuning performance and model generalization in medical video-based machine learning.}
    \label{data_charac}
\end{figure*}

\enluminure{U}{ltrasound (US)} imaging is one of the most widely used modalities in clinical practice~\citep{10981027}, valued for its real-time capability, safety, and cost-effectiveness~\citep{jia2022ultrasound,boumeridja2025enhancing,bellomo2023feasibility}.
Accurate segmentation of anatomical structures in ultrasound videos is essential for applications such as fetal heart assessment~\citep{sobhaninia2019fetal}, liver boundary delineation~\citep{li2025liver, atabansi2025ict}, and cardiac chamber analysis~\citep{leclerc2019camus}.
Unlike static modalities such as CT or MRI, ultrasound video segmentation presents additional challenges due to temporal dynamics, motion artifacts, and variability in acquisition conditions~\citep{yao2025self}.
These factors make supervised segmentation methods heavily dependent on large-scale annotated datasets, which are costly and time-consuming to obtain~\citep{ferreira2025self} .
In addition, conventional deep learning models often struggle to generalize across different organs, views, and acquisition settings, limiting their clinical scalability.

Recent progress in vision foundation models, particularly the Segment Anything Model (SAM)~\citep{kirillov2023segment} and its successor SAM2~\citep{ravi2024sam}, has introduced powerful zero-shot and prompt-guided segmentation capabilities.
The SAM2 architecture integrates temporal memory with flexible prompt encoding, which together enable coherent object tracking across frames and versatile interaction across tasks. These features make SAM2 a promising framework for video segmentation, offering both reduced annotation effort and strong adaptability to diverse clinical scenarios.
However, when applied directly to medical data, SAM2 suffers from pronounced domain shift~\citep{he2408short,chukwujindu2025improving,shen2024interactive,sengupta2025sam,yu2024novel,shen2025performance,mansoori2024polyp,yu2024sam,ma2408segment,dong2024segment,kazemi2025semi}, as the characteristics of medical images and videos differ substantially from those of natural scenes.
This gap underscores the need for systematic strategies to adapt SAM2 to medical imaging, particularly ultrasound video segmentation, where temporal incoherence and imaging variability pose unique challenges.

Efforts to adapt SAM2 for medical image and video segmentation can be broadly categorized into four methodological directions, two or more of which are often combined in individual studies. First, \textit{modality adaptation} through fine-tuning or training-free strategies enables SAM2 to handle domain-specific data such as CT, MRI, and ultrasound \citep{yan2024biomedicalsam2segment, li2024adapting,mansoori2025advancements,hoyer2025clinicalutilityfoundationsegmentation,he2025few,zhu2024medical,ma2025medsam2segment3dmedical,yildiz2024sam,chen2025sam,xiong2408sam2}. Notable models in this context include MedSAM2 \citep{ma2025medsam2segment3dmedical} and Medical-SAM2 \citep{zhu2024medical}, which fine-tune all the SAM2 parameters on curated massive medical image datasets. Second, \textit{prompt engineering} aims to improve segmentation accuracy by designing automatic or task-specific prompts \citep{yu2025crisp,podvin2025samusa,zhang2024path,gutierrez2025prompt,xinyi2025proxy,zu2025rethinking,zhao2025retrieval,xie2025rfmedsam,wei2025self,mansoori2025self}. Third, \textit{parameter-efficient fine-tuning} techniques such as LoRA and adapters allow selective tuning of SAM2 components, significantly reducing computational cost while improving performance on domain-specific tasks\citep{xu2025depthwise,wang2025freqsam2,chen2024sam2,xing2025sam2sgp,wei2025selfpromptingdrivensam2}. Fourth, \textit{architecture refinement} focuses on enhancing SAM2 encoder \citep{huo2025samba, xiong2408sam2, rasaee2025groundingdinoussamtextpromptedmultiorgansegmentation,wahd2025sam2radsegmentation} and memory modules \citep{chen2025accelerating}. Beyond medical imaging, models like EfficientTAM distill SAM2 into lightweight variants suitable for real-time video segmentation on edge devices \citep{xiong2024efficienttam}.

\begin{table*}[t]
\centering
\renewcommand{\arraystretch}{1.2}
\setlength{\tabcolsep}{8pt} 
\caption{Comparison of SAM2 model variants by parameter size and fine-tuning status on ultrasound (US) data. Models are ordered from left to right by decreasing parameter count.}
\label{tab:sam2_models}
\begin{tabular}{|c|c|c|c|c|c|c|c|c|} 
\hline
\textbf{Model} & SAM2-l & SAM2-b & SAM2-s & SAM2-t & MedSAM2 & Medical SAM2 & EFT-t & EFT-s \\
\hline
\textbf{\# Parameters} & 224.4M & 80.8M & 46M & 38.9M & 38.9M & 38.9M & 34M & 18M \\
\hline
\rowcolor{red!10} \textbf{FT on US} & \ding{55} & \ding{55} & \ding{55} & \ding{55} & \cellcolor{green!10}\ding{51} & \cellcolor{green!10}\ding{51} & \ding{55} & \ding{55} \\
\hline
\end{tabular}
\end{table*}


Despite encouraging progress, the optimal fine-tuning strategy for adapting SAM2 to medical video segmentation - especially ultrasound video segmentation - remains unclear. In particular, the influence of data characteristics such as training-set size, video duration, and spatial or temporal augmentations has not been systematically investigated. Unlike the well-curated natural video datasets used for SAM2 training, ultrasound videos exhibit substantial variability in acquisition process and anatomical appearance. As illustrated in Figure~\ref{data_charac}, both inter-dataset (domain shift, uneven dataset scale, acquisition parameter variations) and intra-dataset (variations in clip length, motion diversity, frame rate) heterogeneity of ultrasound videos can substantially affect adaptation and generalization. For instance, the data imbalance across clinical tasks raises questions about how fine-tuning performance scales with dataset size. Temporal structure also presents an open question during fine-tuning: While longer videos capture richer temporal dynamics, incorporating shorter clips during fine-tuning enhances data diversity at the cost of reduced utilization of SAM2’s memory and tracking mechanisms.
Whether short sequences contribute effectively to model adaptation is still unclear. A systematic, data-centric investigation is thus essential to reveal how dataset composition and temporal context shape SAM2 fine-tuning, guiding the development of robust and efficient adaptation pipelines for ultrasound video segmentation.


To address these gaps, we present a systematic study on fine‑tuning SAM2 for ultrasound video segmentation. From a data‑centric perspective, our investigation focuses on three key questions:
(i) \textbf{How does fine-tuning performance scale with dataset size?} We evaluate scaling behavior under four settings: zero‑shot (no training data), and fine‑tuning with 25\%, 50\%, and 100\% of the available training set.
(ii) \textbf{How does video duration affect adaptation?} We partition the dataset into long and short videos using a frame‑count threshold of $N=8$ and further conduct a fine‑grained analysis by incrementally adding short videos of varying lengths ($N = 1-7$) during finetuning to assess their contributions.
(iii) \textbf{Can spatial and temporal augmentations boost performance?} We propose a set of ultrasound‑specific temporal and spatial augmentation strategies and systematically compare them against two baselines: the original SAM2 augmentation pipeline and a no‑augmentation setting. This comparison aims to quantify the contribution of domain-specific augmentations to model generalization and robustness.

Ultrasound videos present substantial domain variation across datasets, which can markedly influence the fine-tuning behavior and generalization ability of foundation models. To systematically examine this effect, we design a set of training paradigms representing common data-adaptation scenarios in medical imaging. Specifically, we compare three strategies: (i) \textbf{task-specific fine-tuning}, which adapts the model to a single target dataset; (ii) \textbf{intermediate adaptation schemes} (\emph{midtraining} and \emph{midtraining–finetuning}) that incorporate additional ultrasound datasets prior to target-task adaptation; and (iii) \textbf{joint training}, where multiple target datasets are fine-tuned simultaneously to encourage cross-dataset generalization. 

Beyond training strategies, factors related to \textbf{model architecture} and \textbf{initialization} remain insufficiently explored in SAM2 adaptation. Most prior studies rely on a single backbone, leaving open how architectural diversity and pretrained weights influence downstream transfer. To investigate this, we evaluate five representative backbone–weight configurations, including \textbf{SAM2} and its \textbf{medical} and \textbf{efficiency-oriented variants}, across three prompting modes commonly used in interactive segmentation: single-click (\textbf{Click}), bounding box (\textbf{BBox}), and dense mask (\textbf{Mask}). Each prompting mode is applied consistently during both training and inference.


Overall, this study provides a data-centric analysis of SAM2 fine-tuning. By examining how data composition, training strategies, and model initialization interact, we aim to provide empirical insights for developing efficient and robust ultrasound segmentation frameworks grounded in foundation model adaptation.

\begin{table*}[htbp]
\centering
\caption{Dataset summary with structures listed individually. Green rows: used for task-specific fine-tuning (FT), midtraining (MT), midtraining-finetuning (MT=FT) and joint training (JT). Blue rows: used for MT/MT-FT.}
\label{tab:dataset}
\setlength{\tabcolsep}{4.0pt}
\renewcommand\theadfont{\normalsize\bfseries}
\renewcommand{\arraystretch}{1.2} 
\begin{threeparttable}
\small
\begin{tabularx}{\textwidth}{@{} l c c l c c l @{}}
\toprule
\thead{Dataset} & \thead{\# Videos} & \thead{\# Images} & \thead{Structure} & \thead{\# Image--mask pairs} & \thead{Total pairs} & \thead{Usage} \\
\midrule
\rowcolor{green!10}
FetalHeart 3VT & 637 & 2120 & Aortic Arch & 1779 & 9837 & \textbf{FT/MT/MT-FT/JT} \\
\rowcolor{green!10}
 &  &  & Ductus Arch & 2100 &  &  \\
\rowcolor{green!10}
 &  &  & SVC & 2079 &  &  \\
\rowcolor{green!10}
 &  &  & Spine & 1976 &  &  \\
\rowcolor{green!10}
 &  &  & Trachea & 1903 &  &  \\
\rowcolor{green!10}
4-Chamber & 1114 & 5098 & LA & 2873 & 14365 & \textbf{FT/MT/MT-FT/JT} \\
\rowcolor{green!10}
 &  &  & LV & 2873 &  &  \\
\rowcolor{green!10}
 &  &  & RA & 2873 &  &  \\
\rowcolor{green!10}
 &  &  & RV & 2873 &  &  \\
\rowcolor{green!10}
 &  &  & Spine & 2873 &  &  \\
\rowcolor{green!10}
Liver & 494 & 4733 & Liver & 4060 & 4060 & \textbf{FT/MT/MT-FT/JT} \\
\rowcolor{blue!8}
SonoAVC & 213 & 26582 & Ovary & 10541 & 171212 & \textbf{MT/MT-FT} \\
\rowcolor{blue!8}
 &  &  & Follicles & 160671 &  &  \\
\rowcolor{blue!8}
Pelvic Floor & -- & 680 & Levator ani & 680 & 1360 & \textbf{MT/MT-FT} \\
\rowcolor{blue!8}
 &  &  & Symphysis pubis & 680 &  &  \\
\rowcolor{blue!8}
Common Bile Duct & -- & 67 & CBD & 67 & 67 & \textbf{MT/MT-FT} \\
\rowcolor{blue!8}
Kidney & -- & 898 & Kidney & 591 & 591 & \textbf{MT/MT-FT} \\
\bottomrule
\end{tabularx}
\begin{tablenotes}[flushleft]
\footnotesize
\item \textit{Notes.} Repeated fields within each dataset block are intentionally left blank after the first line. Dashes (--) denote values not reported.
\end{tablenotes}
\end{threeparttable}
\end{table*}




\begin{figure*}[t]
    \includegraphics[width=1.0\textwidth]{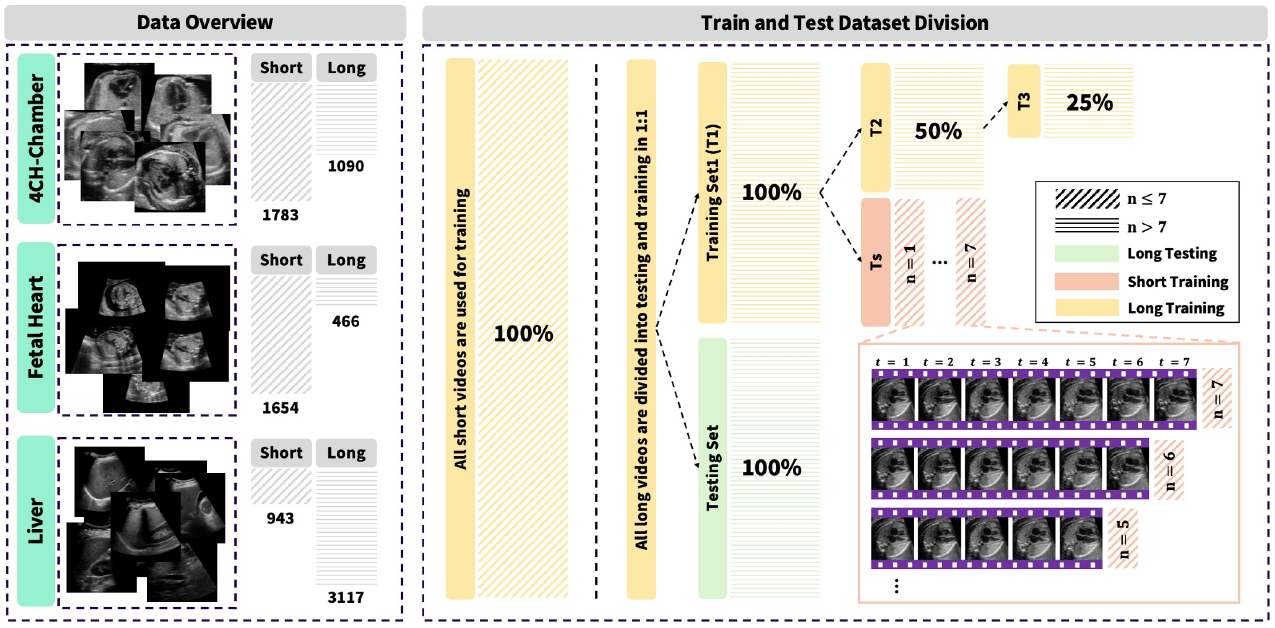}
    \caption{Dataset overview and partition scheme. Left: overview of the three core target datasets, showing frame counts for short and long videos. Right: illustration of the training–testing split, construction of multi-scale training subsets, and generation of the synthetic balanced short-video training set.}
    \label{fig:dataset}
\end{figure*}

\section{Methods}
\label{sec:methods}
This section contains five main components: \textbf{Section~\ref{subsec:models}} introduces the core segmentation models evaluated in this work.
\textbf{Section~\ref{subsec:datasets}} describes the data preprocessing and data division for training and evaluation.  
\textbf{Section~\ref{subsec:data_variation}} details the dataset construction strategy designed to explore different types of data variation, with an emphasis on training-set scale and temporal context stratification.  
\textbf{Section~\ref{subsec:strategy}} presents the adaptation regimes.
\textbf{Section~\ref{subsec:aug}} presents the ultrasound-specific spatial and temporal augmentation strategies we proposed to enhance model generalization and robustness.

\begin{figure*}[h]
    \includegraphics[width=1\textwidth]{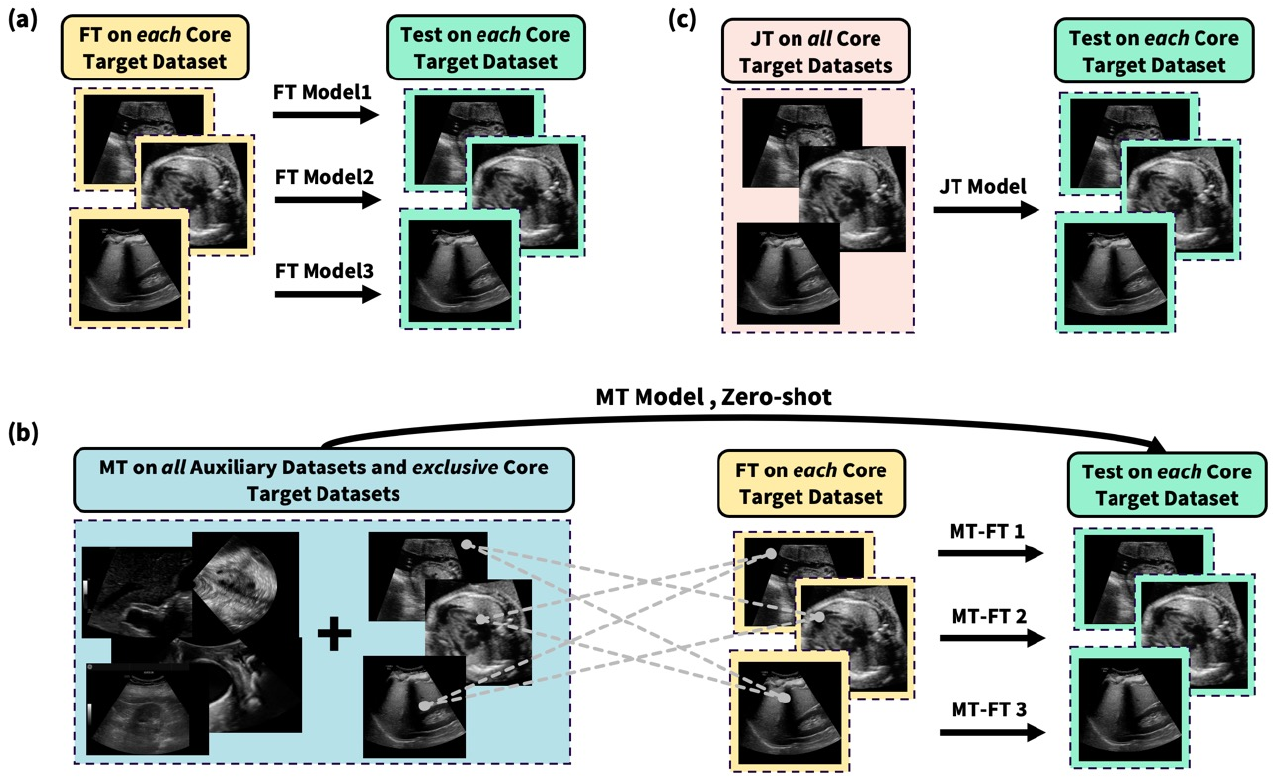}
    \caption{
    Illustration of adaptation regimes in this study. 
    (a) Task-specific fine-tuning (FT): each model is fine-tuned and tested on a specific core target dataset (with training and testing dataset split). 
    (b) Midtraining (MT) and midtraining–fine-tuning (MT-FT): for each \textit{Core Target Dataset}, MT is performed on all \textit{Auxiliary} and the rest of \textit{Core Target Datasets}. 
    During MT-FT, Gray dashed lines indicate the corresponding training \textit{Core Target Datasets} for each testing one. 
    (c) Joint Training (JT): model is trained on all \textit{Core Target Datasets} simultaneously.
    }

    \label{fig:adaptation}
\end{figure*}

\subsection{Model Architectures}
\label{subsec:models}

To comprehensively assess segmentation performance across model configurations, multiple widely-used SAM2 variants are included in this study. Table~\ref{tab:sam2_models} lists their parameter scales and indicates whether each has been fine-tuned on ultrasound data. These models fall into 3 categories defined by their backbone design and pretraining datasets:

\noindent \textbf{SAM2 group:} This group includes all models proposed in the SAM2 paper~\citep{ravi2024sam}: SAM2.1-Hiera-Large (SAM2-l, 224.4M), SAM2.1-Hiera-Base (SAM2-b, 80.8M), SAM2.1-Hiera-Small (SAM2-s, 46M), and SAM2.1-Hiera-Tiny (SAM2-t, 38.9M).
SAM2 extends the image-only SAM~\citep{kirillov2023segment} to video segmentation by combining a Hiera-pretrained hierarchical ViT encoder~\citep{ryali2023hiera} with a temporal memory mechanism that maintains consistency across frames. Each frame is encoded independently, and a transformer decoder predicts masks conditioned on three inputs: (i) current frame features, (ii) user prompts, and (iii) a memory bank summarizing previous frames.
Temporal coherence is achieved through memory attention over compact representations of past masks and features, while a lightweight memory encoder updates the bank iteratively. Prompts are embedded with positional encodings and learned foreground/background tokens, supporting flexible inputs such as clicks, boxes, or masks. This design preserves SAM’s interactivity while ensuring temporal stability, making SAM2 models widely adopted as backbones in video segmentation foundation models.

\noindent \textbf{Medical group:} This group includes SAM2 variants further fine-tuned on medical data: 
MedSAM2~\citep{ma2025medsam2segment3dmedical} (38.9M) 
and Medical SAM2~\citep{zhu2024medical} (38.9M). 
Both models start from SAM2-t checkpoints and are adapted on curated medical datasets spanning CT, MRI, and ultrasound modalities. 
MedSAM2 largely follows SAM2’s training setup but differs in two aspects: 
it uses an input resolution of $512 \times 512$ instead of $1024 \times 1024$, 
and it restricts prompting to bounding boxes. 
In contrast, Medical SAM2 retains SAM2’s prompting mix and learning-rate schedule but introduces a self-sorting memory bank that dynamically selects informative memory entries 
based on confidence and dissimilarity rather than recency. 
Another distinction is that MedSAM2 is fine-tuned on both 3D volumes and 2D videos, 
whereas Medical SAM2 focuses mainly on 3D volumes. 
In this study, the Medical group is used to assess how medical-domain initialization influences ultrasound video fine-tuning.

\noindent \textbf{EFT group:} The Efficient Track Anything (EFT) family~\citep{xiong2024efficienttam} includes 
EfficientTAM-Small (EFT-s, 34M) and EfficientTAM-Tiny (EFT-t, 18M). 
EFT aims to reduce computational cost and enable real-time video segmentation while retaining a SAM2-style pipeline. 
It replaces the hierarchical Hiera encoder with a plain ViT backbone 
and introduces an optimized memory module that lowers per-frame feature extraction and memory-attention complexity. 
EFT models are trained on the SA-1B (image) and SA-V (video) datasets 
and achieve performance comparable to SAM2 with substantially improved efficiency. 
In this work, the EFT group serves to evaluate the effectiveness of lightweight foundation-model initialization 
for ultrasound video fine-tuning.

\subsection{Datasets}
\label{subsec:datasets}

In this study, we curate an internal ultrasound dataset suite that spans multiple anatomical regions and segmentation targets to rigorously evaluate SAM2 adaptation under different training regimes. As listed in Table~\ref{tab:dataset}, the suite comprises three \textit{core target datasets}—used for task-specific fine-tuning (FT), intermediate pretraining (MT/MT-FT), and joint training (JT)—and four \textit{auxiliary datasets} used exclusively for MT and MT-FT. Across datasets, we use the number of image-mask pairs as the primary supervision budget and the total pairs per dataset as the annotation scale.

For the \textit{core target datasets}, the \textbf{FetalHeart 3VT} includes 637 videos and 2{,}120 images with five annotated structures (Aortic Arch, Ductus Arch, SVC, Spine, Trachea) comprising 1{,}779, 2{,}100, 2{,}079, 1{,}976, and 1{,}903 image-mask pairs (9{,}837 in total). \textbf{4-Chamber} cardiac dataset comprises 1{,}114 videos and 5{,}098 images with Left Atrium (LA), Left Ventricle (LV), Right Atrium (RA), Right Ventricle (RV) and Spine annotations, each contributing 2{,}873 image-mask pairs to a total of 14{,}365 pairs. \textbf{Liver} contains 494 videos and 4{,}733 images with a single Liver target (total 4{,}060 pairs). 

The \textit{auxiliary datasets} expands intra-modality diversity: \textbf{SonoAVC} provides 213 3D volumes and 26{,}582 images with ovary (10{,}541 pairs) and follicle annotations (160{,}671 pairs), totaling 171{,}212 pairs; \textbf{Pelvic Floor} offers 680 images with Levator ani and Symphysis pubis masks (680 pairs each; total 1{,}360); \textbf{Common Bile Duct} includes 67 images with annotations for the common bile duct (total 67 pairs); \textbf{Kidney} contributes 898 images with kidney segmentation (total 591 pairs).

\noindent\textbf{Data preprocessing:} For all the datasets, the ultrasound region of interest was extracted from DICOM images using the DICOM metadata and converted to PNG format. Segmentation masks for each object were binarized. The images and masks were then resized to square dimensions of 1024 pixels.

\subsection{Exploring Different Types of Data Variation}
\label{subsec:data_variation}

As shown in Figure~\ref{fig:dataset}, we design a dataset construction pipeline to investigate how training-set size and temporal context influence the fine-tuning of SAM2.

\noindent \textbf{Categorization of short and long videos:}  
Each video in the \textit{core target datasets} is partitioned into \emph{short} and \emph{long} clips according to its frame count \(n\). Clips are labeled as \emph{short} when \(n < 8\) and \emph{long} when \(n \geq 8\). The threshold \(n = 8\) is derived from the default SAM2 training configuration that samples eight frames per step. For each dataset, all short clips form a dedicated training pool, denoted as \textit{ST}, used to evaluate the contribution of real-world short videos to fine-tuning. Long clips are evenly split (\(1{:}1\)) into training and testing sets, so that only long clips appear in the test set. This design ensures a consistent temporal evaluation setting where SAM2’s memory and tracking mechanisms can be fully engaged, enabling fair comparison across models. We omit a separate validation set and train all models for a fixed number of epochs (\(E = 70\)), consistent with MedSAM2~\citep{ma2025medsam2segment3dmedical}.

\noindent \textbf{Multi-scale training sets:}  
To study how data scale influences performance, we construct training sets of different sizes (Figure~\ref{fig:dataset}). The full set is denoted as \text{T1}. Two smaller subsets, \text{T2} and \text{T3}, are created by randomly sampling \emph{videos} from \text{T1} to 50\% and 25\% of its original size, respectively, simulating limited data scenarios. Sampling is stratified by frame-count bins to preserve the distribution of video lengths; for instance, if \text{T1} contains four 10-frame videos, two are randomly selected for inclusion in \text{T2}.

\noindent \textbf{Synthetic balanced short-video training sets:}  
To control for frame-count imbalance, we generate a synthetic balanced short-video dataset, \text{Ts}, derived from the remaining portion of \text{T1} after \text{T2} extraction. \text{Ts} comprises seven groups with frame numbers \(n \in \{1, \dots, 7\}\). Videos within each group share the same frame count, while the number of videos is balanced across groups to maintain comparable sample sizes. \text{Ts} is incrementally added to the training data to measure the marginal contribution of short clips to the model fine-tuning.

\subsection{Adaptation Regime}
\label{subsec:strategy}

In this section, we introduce the adaptation regimes used in this study: \textbf{task-specific fine-tuning}, \textbf{midtraining and midtraining‑finetuning}, and \textbf{multi-task joint training}, as shown in Figure.\ref{fig:adaptation}.

\subsubsection{Task-specific Fine-tuning (FT)}
\label{subsec:taskft}

FT~\citep{gu2024build} (Figure.\ref{fig:adaptation}(a)) refers to the
supervised fine-tuning of a foundation model to a specific downstream task or dataset. In this study, we systematically evaluate the FT performance of five SAM2 variants under different training data size. The goal is to understand how data characteristics and model design jointly affect fine-tuning effectiveness.
To this end, we conduct a series of controlled experiments:
(i) \textbf{Section~\ref{subsec:preft}} establishes zero-shot baselines for each SAM2 variant prior to fine-tuning;
(ii) \textbf{Section~\ref{subsec:datascale}} analyzes the effect of training data scale by comparing results on subsets containing 25\%, 50\%, and 100\% of the annotated data;
(iii) \textbf{Section~\ref{subsec:shortvideo}} investigates temporal factors by incorporating short-duration ultrasound clips during fine-tuning to assess how video length and temporal context influence model adaptation.
We employ a full end-to-end fine-tuning strategy, updating all model parameters rather than using parameter-efficient alternatives (e.g., adapters or LoRA), in alignment with recent state-of-the-art SAM2-based medical segmentation frameworks \citep{ma2025medsam2segment3dmedical,yan2024biomedicalsam2segment}.

\subsubsection{Midtraining (MT) and Midtraining–Finetuning (MT–FT)}
\label{subsec:midtraining}

MT (Figure.\ref{fig:adaptation}(b)) is a common strategy for adapting foundation models to new domains by leveraging auxiliary datasets from the same imaging modality but distinct from the target task. It enables the model to acquire modality-specific representations before task-level fine-tuning (FT). This paradigm has been adopted in medical foundation models such as MedSAM2~\citep{ma2025medsam2segment3dmedical} and Medical SAM2~\citep{zhu2024medical} to strengthen domain alignment. Building on this idea, the MT-FT paradigm performs MT followed by FT on the target dataset (Figure.\ref{fig:adaptation}(b)), combining broad modality adaptation with task-specific optimization. This two-stage process aims to combine the broad modality adaptation of MT with the task-specific optimization of FT.


\begin{figure*}[t]
    \includegraphics[width=1.0\textwidth]{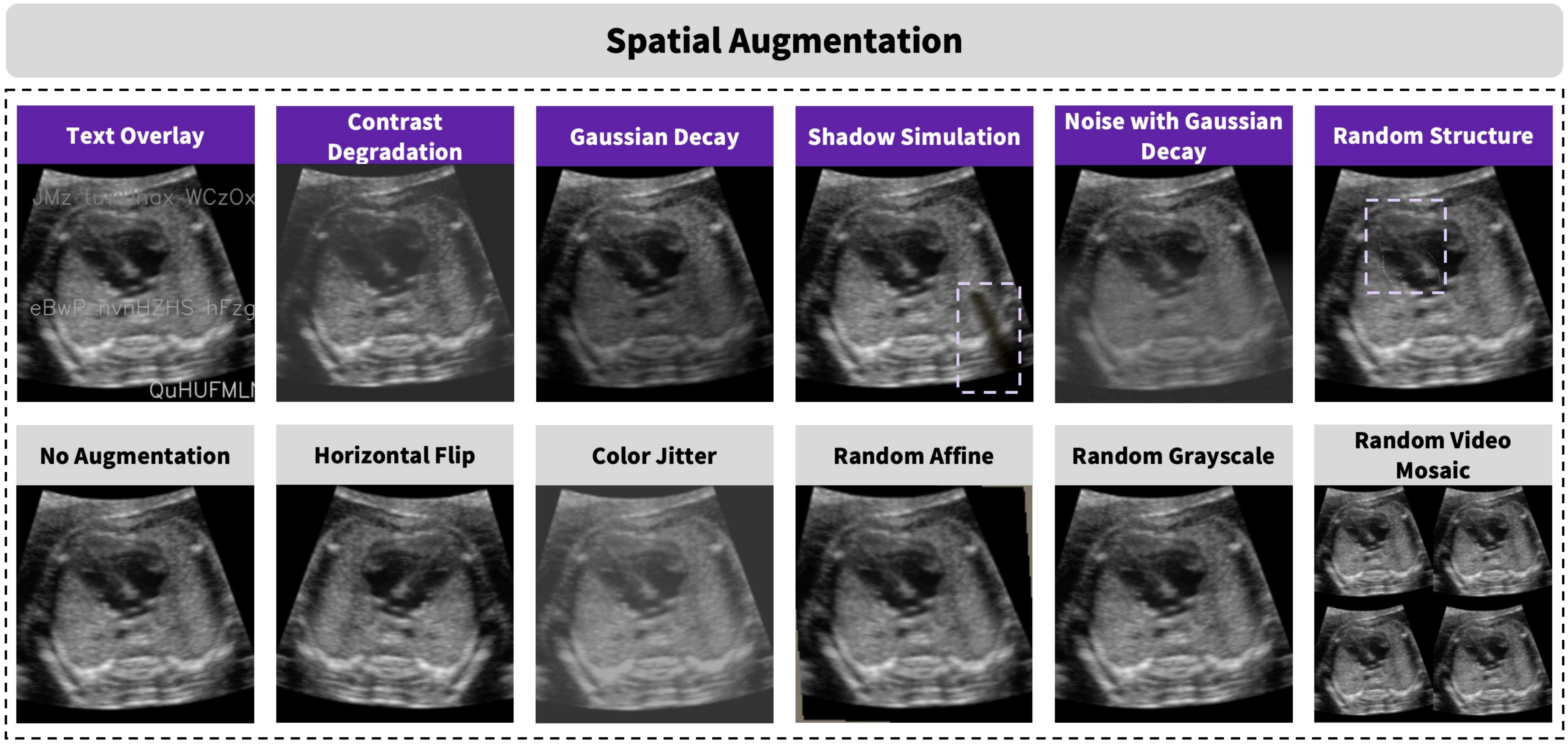}
    \caption{Visualization of the spatial augmentation strategies. The first row (purple) shows augmentations proposed in this study, and the second row (gray) shows the original image augmentations along with the SAM2-default. Dash-lined boxes highlight structural changes in the augmented results.}
    \label{fig:spatialaugs}
\end{figure*}

In adapting SAM~\citep{kirillov2023segment} to medical imaging, \citet{gu2024build} examined both MT and MT–FT (termed “task-expansion finetuning” in their work). They found that MT–FT offered only marginal gains over direct FT while increasing training cost. Our study extends this line of investigation to video segmentation in three major ways. First, we explicitly separate MT from MT–FT and exclude target-task data during MT, providing a clearer view of each adaptation stage. Second, we focus on ultrasound video segmentation rather than static MRI, introducing complementary insights into dynamic and noise-prone imaging. Finally, we emphasize the role of data scale, analyzing how MT and MT–FT behave under different amounts of labeled data to reflect both data-rich and limited-data clinical conditions. 

We systematically evaluate how MT influences zero-shot segmentation on ultrasound videos and how it influence subsequent MT–FT across varying data scales and temporal configurations. Detailed experiments are presented in \textbf{Section~\ref{subsec:midtrainzeroshot}}.

\subsubsection{Multi-Task Joint Training (JT)}
\label{subsec:jointtraining}

JT provides an alternative to FT when multiple annotated datasets are available. Instead of training separate models for each dataset, JT learns from all target datasets simultaneously, encouraging the development of shared representations that enhance cross-domain generalization. A single JT-trained model can thus handle multiple related tasks without further adaptation. 

Despite its appeal, JT remains challenging. Variations in dataset distribution can cause domain conflicts and therefore compromise per-task accuracy. Achieving a balance between generalization across datasets and specialization within each specific task remains a central difficulty. Furthermore, the impact of dataset scale, model capacity, and prompting strategies on JT performance has not been systematically explored in the context of foundation model adaptation.

In \textbf{Section.\ref{subsec:jointtrain}}, we evaluate the effectiveness of JT for ultrasound video segmentation. We compare JT with FT across different data scales, architectures, and prompting modalities, highlighting the trade-offs between accuracy, robustness, and computational efficiency. Our goal is to clarify the practical value of JT for clinical ultrasound applications.

\begin{figure}[t]
    \includegraphics[width=0.5\textwidth]{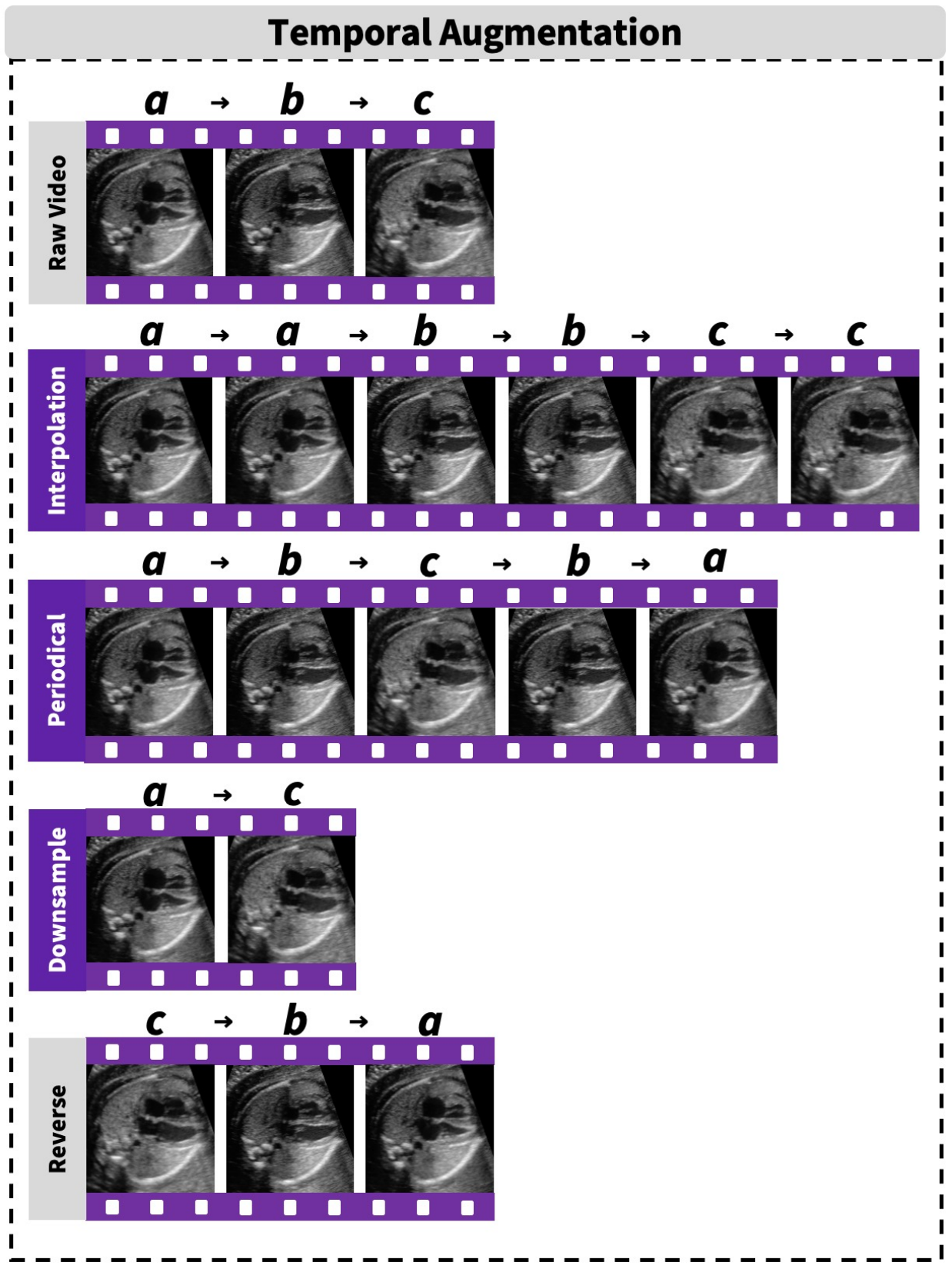}
    \caption{Visualization of the temporal augmentation strategies. The first and last row (gray) shows the SAM2-default augmentations along with the original video. The second to fourth row (purple) shows augmentations proposed in this study. \textit{a}, \textit{b}, \textit{c} denote frame IDs highlighting changes in frame sequence.}
    \label{fig:tempaugs}
\end{figure}

\subsection{Augmentation Techniques}
\label{subsec:aug}

Ultrasound video segmentation is particularly challenging due to its high spatial and temporal variability. 
To mitigate variability in natural images, SAM2 adopts several generic augmentation strategies that enhance prompt sensitivity and mask accuracy. 
These default augmentations have also been adopted in medical adaptations such as MedSAM2~\citep{ma2025medsam2segment3dmedical}. 
However, these augmentations are not tailored to the unique characteristics of ultrasound imaging, where anatomical variability and acquisition artifacts are pronounced. As a result, augmentation strategies specialized for ultrasound remain largely unexplored.

To address this gap, we introduce a set of ultrasound-specific spatial and temporal augmentations that better capture the physical and anatomical variability inherent in ultrasound imaging. 
These augmentations are systematically benchmarked against both the default SAM2 augmentations and a no-augmentation baseline. 
The proposed augmentations fall into two categories:
First, we design three \textbf{temporal augmentations (Figure~\ref{fig:tempaugs})}—\textit{Interpolation (Interp)}, \textit{Downsampling (Sample)}, and \textit{Periodical Sampling (Period)}—to simulate variations in frame rate and temporal continuity. 
Second, six \textbf{spatial augmentations (Figure~\ref{fig:spatialaugs})} are developed to mimic clinically relevant artifacts and signal degradations: 
\textit{Gaussian Decay (GD)} and \textit{Noise with Gaussian Decay (NG)} replicate depth-dependent attenuation and system noise; 
\textit{Random Structure Injection (RS)} and \textit{Text Overlay (TO)} simulate spurious structures and acquisition artifacts; 
\textit{Contrast Degradation (CD)} and \textit{Shadow Simulation (SS)} emulate probe-related contrast loss and rib shadows commonly observed in ultrasound. 
Detailed formulations and implementation settings for each augmentation are provided in \textbf{Appendix}.

For comparison, we also evaluate SAM2 built-in augmentations: \textit{Time Reverse (Reverse)}, \textit{Color Jitter (CJ)}, \textit{Horizontal Flip (HF)}, \textit{Random Affine (RA)}, and \textit{Random Video Mosaic (RV)}, as well as the default SAM2 augmentation pipeline (\textit{DEF}). It is worth noting that among these SAM2 built-in augmentations, \textit{Color Jitter (CJ)} has no effect on the saturation or hue of grayscale images, and thus primarily alters contrast and brightness. In Figure~\ref{fig:spatialaugs}, we visualize \textit{Color Jitter (CJ)} with an exaggerated brightness and contrast variation range ($\pm0.8$) to better illustrate its impact; however, in our experiments, we adopt the default range specified by SAM2. Since \textit{Random Grayscale (RG)} has no effect on grayscale inputs, experiments involving it were omitted.

In \textbf{Section~\ref{subsec:s_aug_train}}, we systematically evaluate all augmentation schemes to quantify their contributions to model robustness and performance. Rather than seeking a universally optimal pipeline—whose effectiveness inevitably depends on dataset characteristics—we aim to establish a clear benchmark by comparing different augmentation categories for SAM2 fine-tuning in ultrasound video segmentation.

\begin{figure*}[h]
    \includegraphics[width=1.0\textwidth]{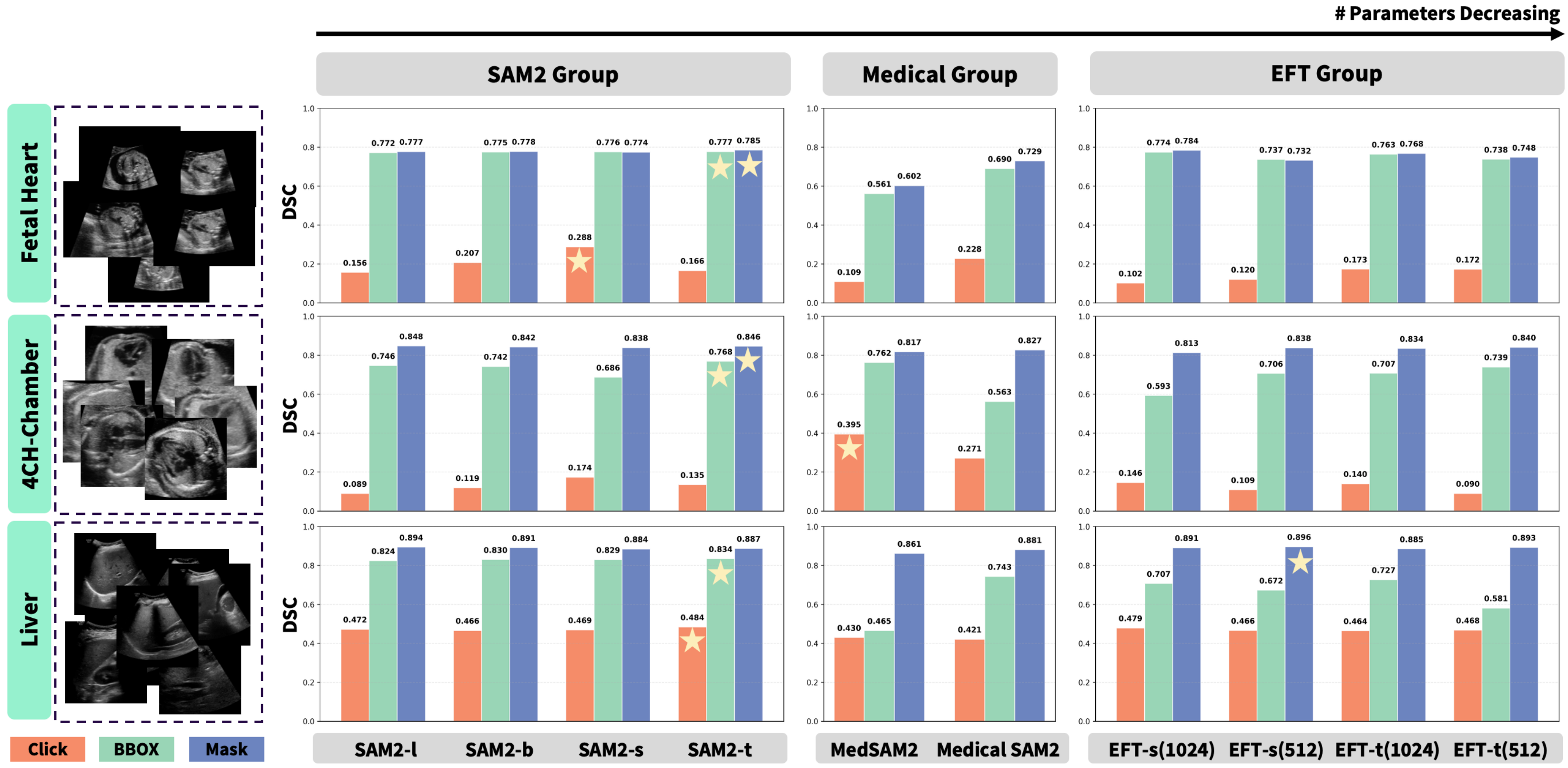}
    \caption{Zero-shot comparison of SAM2 model variants on the \textbf{core target datasets}. DSC performance is shown for three anatomical regions using Click (orange), BBOX (green), and Mask (blue) prompts. Models are grouped by architecture and fine-tuning status. Bright stars highlight the best performance model for each prompt in each dataset.}
    \label{before_ft}
\end{figure*}

\section{Experiments}

In this section, we conduct comprehensive experiments to analyze SAM2 fine-tuning from a data-centric perspective, systematically varying training strategies, model architectures, and initialization schemes. Based on previous works~\citep{ravi2024sam,yan2024biomedicalsam2segment, zhu2024medical}, we adopt full-model fine-tuning with a two-tier learning-rate policy: a lower rate for the image encoder to preserve pre-learned visual representations ($3.0\times 10^{-6}$), and a higher rate for the remaining modules ($5.0\times 10^{-6}$) to adapt to medical-domain characteristics. Mask prediction is optimized using a weighted sum of focal loss and soft Dice loss (weight ratio $20{:}1$), and parameters are updated with the AdamW optimizer~\citep{kingma2014adam} ($\beta_1=0.9$, $\beta_2=0.999$). We employ linear warm-up followed by cosine decay with a minimum learning-rate multiplier of $0.01$. To emulate interactive segmentation in videos, we randomly sample 8-frame sequences and inject up to two corrective interactions per sequence (always including the first frame). Prompts are synthesized from ground-truth masks and current model predictions. The initial prompt for each sequence is sampled with the following probabilities: a ground-truth mask ($50\%$), a single positive point within the mask ($25\%$), or a tight bounding box around the object ($25\%$). Similar to MedSAM2~\citep{ma2025medsam2segment3dmedical}, each model is trained for 70 epochs on a single NVIDIA H100 GPU. Unless otherwise specified, all training reported in this section follow the configuration above. In all the experiments, we use Dice Similarity Coefficient (DSC) scores for evaluation of the performance and average over the last 10 epochs to ensure stable performance. During inference, We use a semi-supervised video object segmentation setting, in which the first frame is prompted and predictions are temporally propagated across the video. The final performance is reported as the mean Dice score averaged over all tracked frames.


\subsection{Zero-Shot Comparison Across SAM2 Variants}

\label{subsec:preft}
\begin{figure*}[t]
    \includegraphics[width=1.0\textwidth]{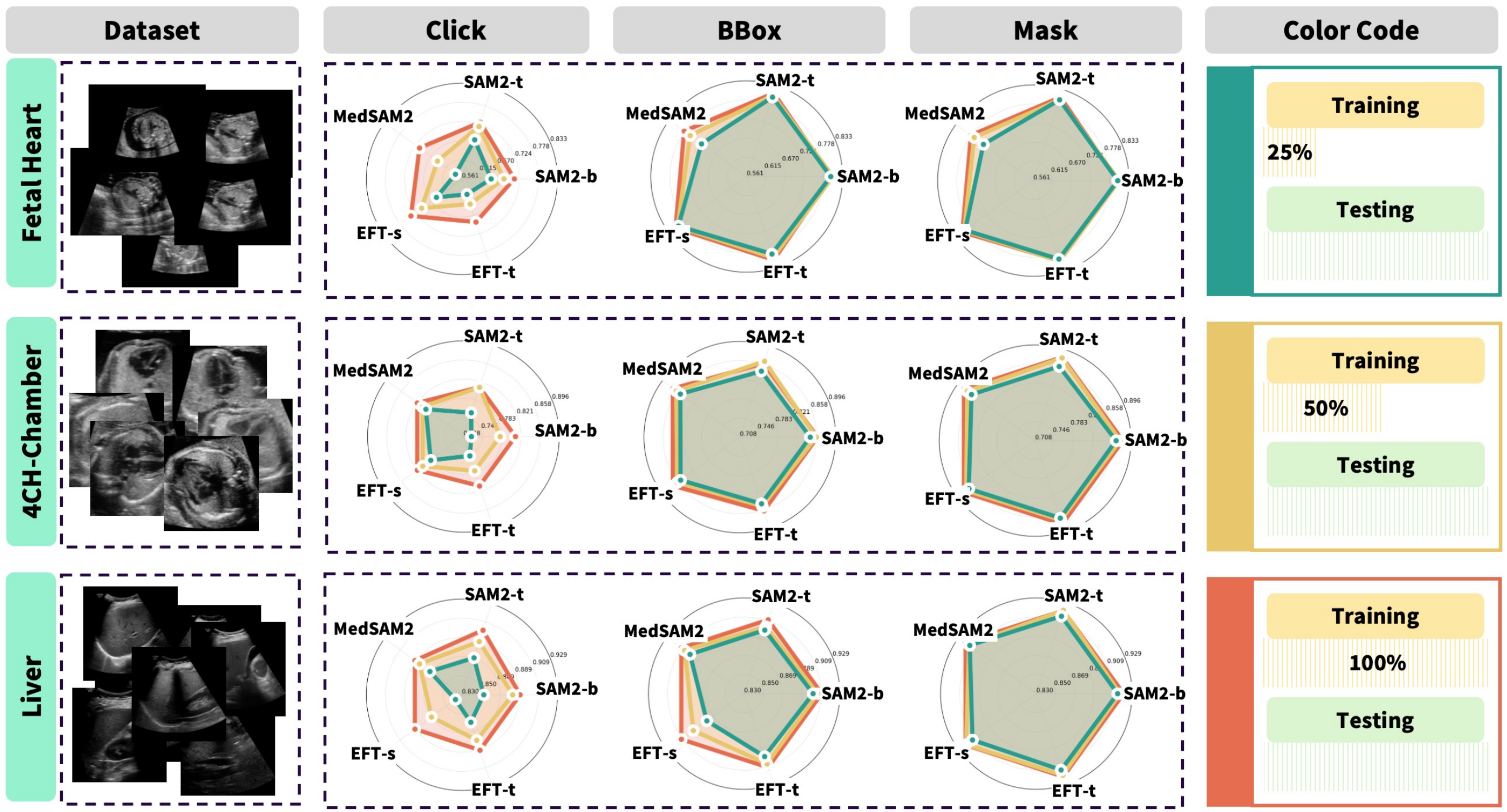}
    \caption{Radar chart comparison of model performance across prompt types and data scales. Color contours represent fine-tuning with 25\% (green), 50\% (yellow), and 100\% (red) of training data.}
    \label{fig:data_scaling}
\end{figure*}

To evaluate the zero-shot segmentation capabilities of different SAM2 model variants on ultrasound videos, we conducted a comparative analysis across multiple architectures and initialization strategies. 

\subsubsection{Implementation Details}

As illustrated in Figure~\ref{before_ft}, the evaluated models are grouped into three categories based on their backbone architecture and pretraining data: the \text{SAM2 Group} (SAM2-l, SAM2-b, SAM2-s, SAM2-t), the \text{Medical Group} (MedSAM2, Medical SAM2), and the \text{EFT Group} (EFT-s/t with varying input resolutions). Models are ordered from left to right by decreasing parameter size. Each row reports the Dice Similarity Coefficient (DSC) on a specific \text{core target dataset}. Results from the three prompting modes—Click, BBox, and Mask—are color-coded in orange, green, and blue respectively for visual clarity. Bright stars indicate the top-performing model for each prompting mode within each dataset.

\subsubsection{Results} 

\noindent\textbf{Model size vs. zero-shot performance:} Prior studies~\citep{zhang2025sam2survey, zhao2025sam2object} have reported clear scaling trends in zero-shot segmentation on natural videos. In contrast, our results reveal that such trends do not consistently hold in the medical domain. Within the SAM2 group, smaller variants (SAM2-s, 46M; SAM2-t, 38.9M) generally outperform larger ones (SAM2-b, 80.8M; SAM2-l, 224.4M). Remarkably, SAM2-t outperforms SAM2-l across nearly all datasets and prompting modes, with a single exception—the Liver dataset under Mask prompting, where SAM2-l performs slightly better.

A similar observation emerges in the EFT group. The lightweight EFT-t (18M) attains performance comparable to or exceeding EFT-s (34M). For example, EFT-t surpasses EFT-s in the Fetal Heart dataset under Click prompting, and in the 4-Chamber dataset under both BBox and Mask modes. In the Liver dataset, EFT-t (1024) achieves the highest BBox score, while EFT-t (512) performs best under Click prompting.

Across all configurations, SAM2-t consistently delivers the strongest results across different datasets and prompting strategies. It achieves the highest performance under Mask and BBox prompting in both the Fetal Heart and 4-Chamber datasets, and also attains the best Click-prompting result in the Liver dataset. SAM2-s performs best for Click prompting in the 4-Chamber dataset, while MedSAM2 (38.9M) achieves comparable Click performance under certain settings. Notably, the lightweight EFT-s (18M, $512\times512$ input) yields the best Mask-prompting accuracy in the Liver dataset. In contrast, neither SAM2-l nor SAM2-b achieves the top score in any dataset–prompt combination.
Overall, these results suggest that scaling laws observed in natural video segmentation may not directly transfer to ultrasound video analysis. Smaller models can deliver competitive—or even superior—zero-shot performance. 

\noindent\textbf{Initialization vs. zero-shot performance:} Previous studies~\citep{zhu2024medical,MedSAM,ma2025medsam2segment3dmedical} have reported clear benefits of medical-domain initialization over original SAM2 weights for zero-shot medical segmentation. In contrast, our results reveal the opposite trend. Except for Click prompting in the 4-Chamber dataset, both MedSAM2 and Medical SAM2 consistently underperform the non-medical SAM2-t across all datasets and prompting modes. Moreover, when compared with the EFT group, which is pretrained on natural videos with smaller parameter sizes, the Medical group shows no consistent advantage and even falls behind in several BBox and Mask prompting cases across all three datasets. This finding challenges the common assumption that medical-domain pretraining necessarily yields superior performance. 

\noindent\textbf{Practical Summary:}These findings suggest that the benefits of current medical foundation models may be limited, potentially due to a domain gap between their fine-tuning datasets and the target ultrasound datasets. This highlights the need for more task-aligned fine-tuning strategies in the medical imaging domain.

\subsection{Impact of Data Scale on Task-Specific Fine-tuning (FT)}
\label{subsec:datascale}

\begin{figure*}[!t]
    \includegraphics[width=0.97\textwidth]
    {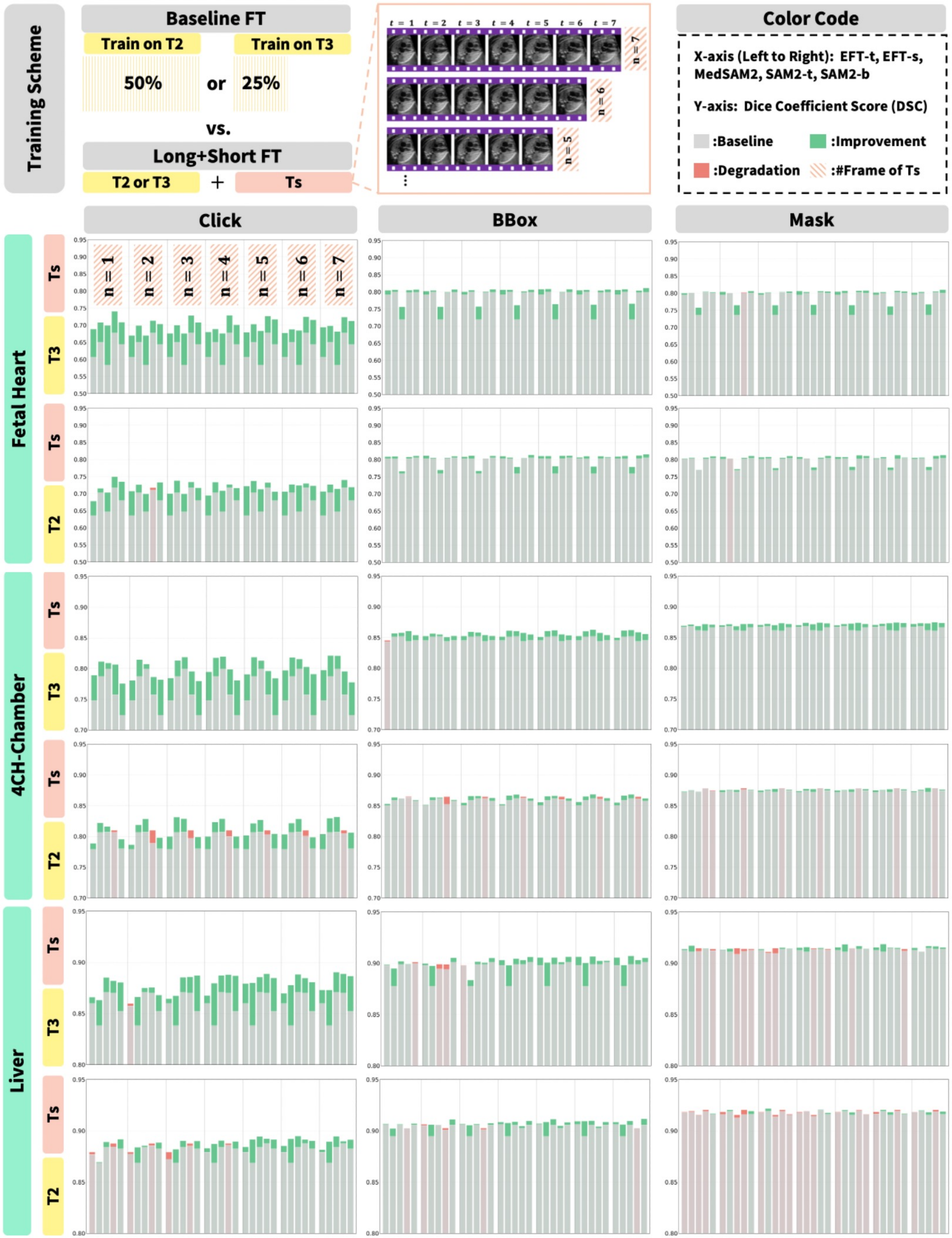}
    \caption{
    Fine-tuning performance after incorporating synthetic balanced short videos under different training scales. 
    }

    \label{fig:long_short}
\end{figure*}

\subsubsection{Implementation Details}

As discussed in \textbf{Section.\ref{subsec:taskft}}, here we perform FT on three \textit{core target datasets} using training scales of 25\% (T3), 50\% (T2), and 100\% (T1) of the available data. Details of dataset construction are provided in Section~\ref{subsec:datasets}. Our experiments cover five representative models drawn from three groups: SAM2-b, SAM2-t, MedSAM2, EFT-s (1024), and EFT-t (1024). These models are selected for both their strong zero-shot performance (Section~\ref{subsec:preft}) and their diversity in architecture (SAM2-based vs. EFT-based), parameter scale (18M–80.8M), and initialization (medical vs. non-medical). This design enables a comprehensive evaluation of FT behavior across heterogeneous model families.

\subsubsection{Results}

Results are visualized in Figure~\ref{fig:data_scaling}, where each row corresponds to a \textit{core target dataset}, and each column represents the DSC under different prompting strategies. Within each radar chart, color contours represent model performance under different data scales, ranging from green (25\%) to yellow (50\%) and red (100\%). This visualization enables direct comparison of how segmentation accuracy varies with data size, model, and prompt type.

\noindent \textbf{Data size vs. performance:} We observe that segmentation accuracy improves consistently with increasing data amount. As the training scale expands, radar contours shift outward, indicating systematic gains across datasets and prompting strategies. The effect is strongest for Click and BBox prompts, while overall performance follows a clear hierarchy: Mask $>$ BBox $>$ Click. Click prompting benefits the most from larger datasets, likely due to its lower initial accuracy, whereas Mask shows only modest gains, likely because its baseline accuracy is already high.

\noindent \textbf{Model size vs. performance.}
The influence of model architecture and initialization diminishes as training data increase. With limited data (25\%, green contours), models exhibit irregular performance patterns, whereas at full-scale training (100\%, red contours), the contours become more uniform and pentagonal, indicating convergence toward similar accuracy regardless of model size or initialization. This suggests that SAM2 FT in the medical domain does not always follow conventional scaling behavior. The trend is consistent with our zero-shot findings (\textbf{Section~\ref{subsec:preft}}) and recent studies on medical foundation model adaptation~\citep{liu2025does,arasteh2025resolution,ma2024segment}, which similarly report that larger models do not necessarily yield better results.

\noindent \textbf{Practical Summary:} These findings provide concrete guidance for model adaptation in medical imaging. First, data scale matters more than architectural complexity or initialization: collecting and curating sufficiently large, high-quality ultrasound datasets should be prioritized over designing ever-larger models. Second, when adequate data are available, lightweight models offer the most practical trade-off between accuracy and efficiency, delivering competitive performance while drastically reducing computational and deployment costs. Together, these insights suggest that progress in medical foundation models will rely less on scaling parameters and more on scaling data.

\subsection{Can short videos contribute to the finetuning?}
\label{subsec:shortvideo}
As mentioned in \textbf{Section~\ref{sec:intro}}, ultrasound videos exhibit substantial variations in acquisition quality and strong imbalance in video lengths within the same dataset (Figure~\ref{data_charac}). While longer videos capture richer temporal dynamics, incorporating shorter clips during fine-tuning increases data diversity and may improve model generalization and overall performance. However, short clips contain too few frames to fully exploit SAM2’s memory and tracking mechanisms. It therefore remains unclear whether shorter sequences meaningfully contribute to model adaptation or whether longer videos alone are sufficient. Clarifying this relationship is critical for developing a robust and data-efficient fine-tuning strategy.

\begin{table*}[htbp]
\centering
\renewcommand{\arraystretch}{1.1} 
\setlength{\tabcolsep}{9pt} 
\caption{Performance comparison of multiple models across datasets and prompt types under different fine-tuning settings: 
\textit{No FT} (no fine-tuning), \textit{T1} (fine-tuned on T1 long videos), and \textit{T1+ST} (fine-tuned on both T1 and real-world short videos).}
\label{tab:ft_longshort}
\begin{tabular}{@{}ll|ccc|ccc|ccc@{}}
\toprule
\textbf{Model} & \textbf{Prompt} & \multicolumn{3}{c|}{\textbf{Fetal Heart}} & \multicolumn{3}{c|}{\textbf{4-Chamber}} & \multicolumn{3}{c}{\textbf{Liver}} \\
              &                 & No FT & T1 & T1+ST & No FT & T1 & T1+ST & No FT & T1 & T1+ST \\
\midrule
\multirow{3}{*}{SAM2-b} & Click & 0.156 & 0.704 & \textbf{0.797} & 0.119 & 0.805 & \textbf{0.841} & 0.466 & 0.890 & \textbf{0.899} \\
       & Box   & 0.772 & 0.804 & \textbf{0.828} & 0.742 & 0.858 & \textbf{0.870} & 0.830 & 0.911 & \textbf{0.915} \\
       & Mask  & 0.777 & 0.804 & \textbf{0.824} & 0.842 & 0.875 & \textbf{0.881} & 0.891 & \textbf{0.921} & \textbf{0.921} \\
\midrule
\multirow{3}{*}{SAM2-t} & Click & 0.166 & 0.725 & \textbf{0.800} & 0.135 & 0.811 & \textbf{0.839} & 0.484 & 0.895 & \textbf{0.903} \\
       & Box   & 0.777 & 0.808 & \textbf{0.827} & 0.769 & 0.863 & \textbf{0.869} & 0.834 & 0.907 & \textbf{0.911} \\
       & Mask  & 0.785 & 0.810 & \textbf{0.820} & 0.846 & 0.879 & \textbf{0.883} & 0.887 & 0.920 & \textbf{0.921} \\
\midrule
\multirow{3}{*}{MedSAM2} & Click & 0.109 & 0.712 & \textbf{0.769} & 0.395 & 0.822 & \textbf{0.848} & 0.430 & 0.894 & \textbf{0.904} \\
        & Box   & 0.561 & 0.778 & \textbf{0.798} & 0.762 & 0.870 & \textbf{0.872} & 0.465 & 0.911 & \textbf{0.913} \\
        & Mask  & 0.602 & 0.777 & \textbf{0.798} & 0.817 & 0.880 & \textbf{0.882} & 0.861 & 0.919 & \textbf{0.921} \\
\midrule
\multirow{3}{*}{EFT-s} & Click & 0.102 & 0.740 & \textbf{0.797} & 0.146 & 0.823 & \textbf{0.840} & 0.479 & 0.887 & \textbf{0.889} \\
      & Box   & 0.775 & 0.811 & \textbf{0.828} & 0.593 & 0.866 & \textbf{0.868} & 0.707 & 0.905 & \textbf{0.911} \\
      & Mask  & 0.784 & 0.812 & \textbf{0.822} & 0.813 & 0.876 & \textbf{0.881} & 0.891 & 0.917 & \textbf{0.920} \\
\midrule
\multirow{3}{*}{EFT-t} & Click & 0.173 & 0.687 & \textbf{0.773} & 0.140 & 0.807 & \textbf{0.822} & 0.464 & \textbf{0.886} & 0.885 \\
      & Box   & 0.764 & 0.808 & \textbf{0.820} & 0.707 & 0.855 & \textbf{0.858} & 0.727 & 0.906 & \textbf{0.909} \\
      & Mask  & 0.768 & 0.804 & \textbf{0.815} & 0.834 & 0.876 & \textbf{0.878} & 0.885 & 0.918 & \textbf{0.919} \\
\bottomrule
\end{tabular}
\end{table*}

\subsubsection{Implementation Details}


We design two experiments to investigate the role of short-duration ultrasound clips in FT.  
(i) To examine how incorporating short clips influences performance under different data scales, models fine-tuned solely on T2 (50\%) and T3 (25\%) serve as baselines. These are compared with counterparts trained on T2 or T3 augmented with synthetic, length-balanced short videos (Ts) containing clips of varying frame counts \(n\) (Figure~\ref{fig:dataset}). 
(ii) To assess the practical impact of short videos in real-world clinical settings, we compare a model FT on T1 with another trained on T1 augmented with real short clips (ST). The results are presented in Table~\ref{tab:ft_longshort}.

\begin{figure*}[t]
    \includegraphics[width=1.0\textwidth]{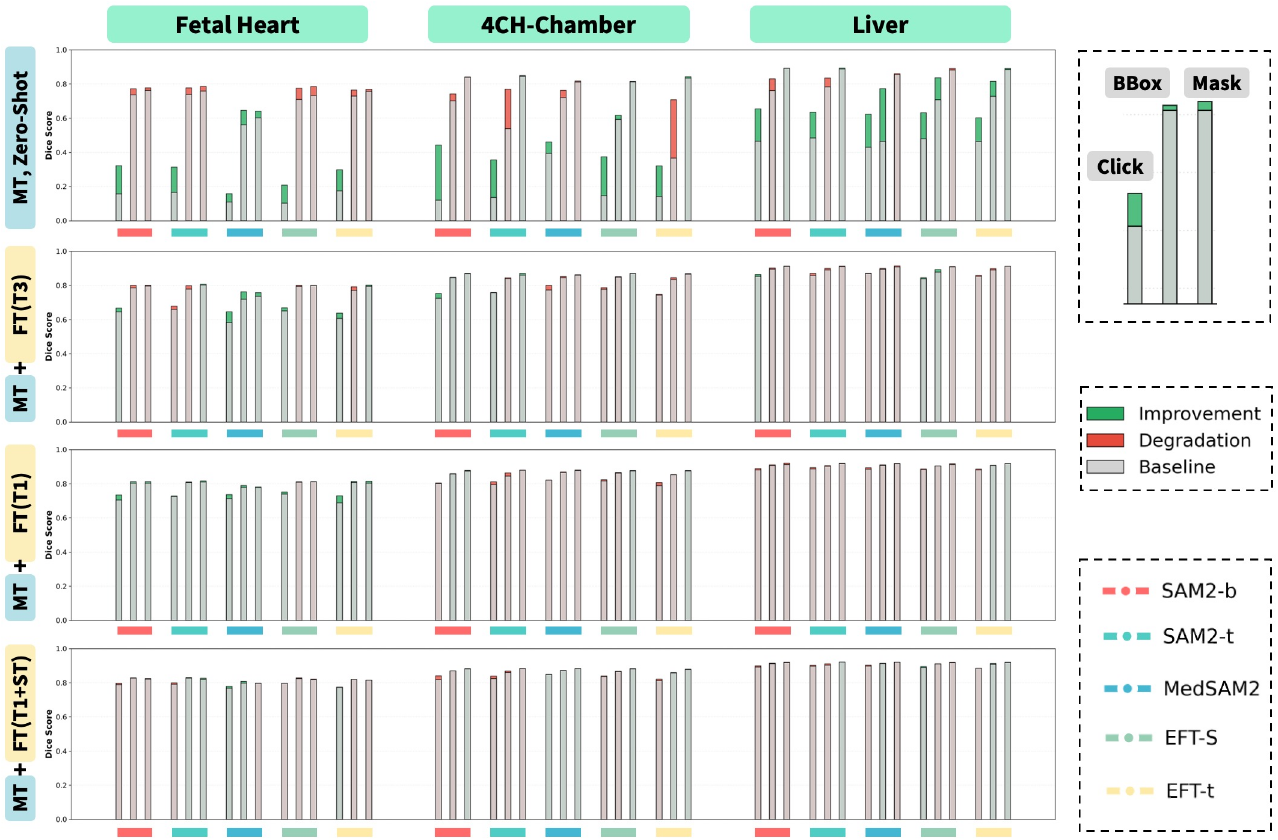}
    \caption{
Midtraining (MT) and midtraining–finetuning (MT–FT) results on the \textit{core target datasets}.
The first row shows the zero-shot performance of the MT model, while the second to fourth rows present MT–FT results on 
\textit{T3} (25\%), \textit{T1} (100\%), and \textit{T1+ST} (combined with real-world short-video training data), respectively.
}

    \label{fig:midtrain}
\end{figure*}

\subsubsection{Results}

Figure~\ref{fig:long_short} presents FT results after incorporating short videos across different training scales. Each row to a specific dataset–scale combination, and each column corresponds to a prompting strategy. Within each block, DSC are arranged to show performance changes as short-video length decreases from \(n=1\) to \(n=7\). Results are reported for five models: EFT-t, EFT-s, MedSAM2, SAM2-t, and SAM2-s, listed in this order from left to right for each combination of prompt, dataset scale and $n$. Gray bars denote baseline FT performance obtained using long videos only (T2 or T3), while improvements and degradations after adding short clips are highlighted in green and red, respectively. To maintain comparability, all DSC plots share consistent value ranges across scales within each dataset.

\noindent\textbf{Contribution of Synthetic Short Videos:}  
As shown in Figure~\ref{fig:long_short}, the prevalence of green regions, which indicate performance gains, shows that incorporating short videos generally improves FT performance across different training scales. A comparison between T3+Ts and T2+Ts further reveals that, for the same short-video length \(n\), the smaller-scale setting (T3+Ts) consistently achieves greater improvements and fewer degradations across all datasets and prompting strategies. This trend suggests that short videos contribute more significantly in limited-data regimes, while their effect becomes less pronounced when sufficient long-video data are already available.

We further observe that the influence of short videos varies across prompting strategies, model architectures, and datasets. The benefit of incorporating short clips gradually decreases from Click to Mask, with Mask prompting under the T2+Ts setting showing the smallest performance change, likely due to its higher baseline accuracy leaving limited room for improvement. Model responses also differ: while most models achieve moderate gains, SAM2-t consistently exhibits performance declines after adding short videos. For example, it shows degradation in the Fetal Heart dataset under (T2+Ts, \(n=2\), Click) and (T3+Ts, \(n=2\), Mask), and similar patterns appear in the 4-Chamber and Liver datasets. To ensure fair comparison, SAM2-t is excluded from subsequent analyses.

The effect of adding short videos also varies across datasets. In the Fetal Heart and 4-Chamber datasets, all models except SAM2-t show consistent performance gains across training ratios and prompting strategies, with only a few minor degradations observed at very small frame numbers (\(n=1\) or \(n=2\)). In the Liver dataset, excluding SAM2-t, occasional degradations appear under T3 and T2 with Click and BBox prompting. Under Mask prompting, improvements dominate in the T3 setting, while in T2 the ratio of improvements to degradations is nearly balanced (13:15, excluding SAM2-t), suggesting that the benefits of short videos are limited. This further supports the earlier finding that the effect of short videos diminishes once sufficient training data are available.

We further investigate how short-clips with different frame counts influence fine-tuning performance. As shown in Figure~\ref{fig:long_short}, only minor differences are observed across frame numbers within each block. This likely results from our construction strategy, where clips are composed of consecutive frames sampled at a rate of 1, leading to high inter-frame similarity. After excluding SAM2-t, degradations appear more frequently in groups with smaller frame numbers, suggesting that longer clips generally offer advantages. Interestingly, the \(n=1\) condition reveals that even single-frame images can noticeably improve performance, likely because their inclusion increases data diversity, which compensates for the lack of temporal information and enhances model generalization.

\noindent\textbf{Contribution of real-world short clips:}  
Table~\ref{tab:ft_longshort} presents an analysis of how short real-world clips contribute to model performance. We compare a baseline model fine-tuned only on T1 with another model fine-tuned on T1 combined with ST. Across all models and prompting strategies, adding ST leads to consistent improvements on the Fetal Heart and 4-Chamber datasets, and noticeably enhances performance in most cases for the Liver dataset. In the few cases where the improvement is limited, T1+ST performs on par with T1 without any noticeable drop. When examining the relative data proportions, both Fetal Heart and 4-Chamber have a higher ST-to-T1 ratio than the Liver dataset. These findings confirm our earlier observation that short clips provide greater benefits when the available training data are limited, while their effect becomes less pronounced once the dataset is relatively sufficient.

\noindent\textbf{Practical summary:}  
Our findings reveal a clear inverse relationship between the effectiveness of short videos and both dataset size and baseline accuracy. From a practical perspective, two key implications emerge. (1) When the training data are limited and baseline performance is low, incorporating short videos yields substantial improvements. (2) When large-scale training data are available and the baseline is already strong, short videos can be omitted to reduce computational cost without sacrificing performance.

\begin{figure*}[t]
    \includegraphics[width=1.0\textwidth]{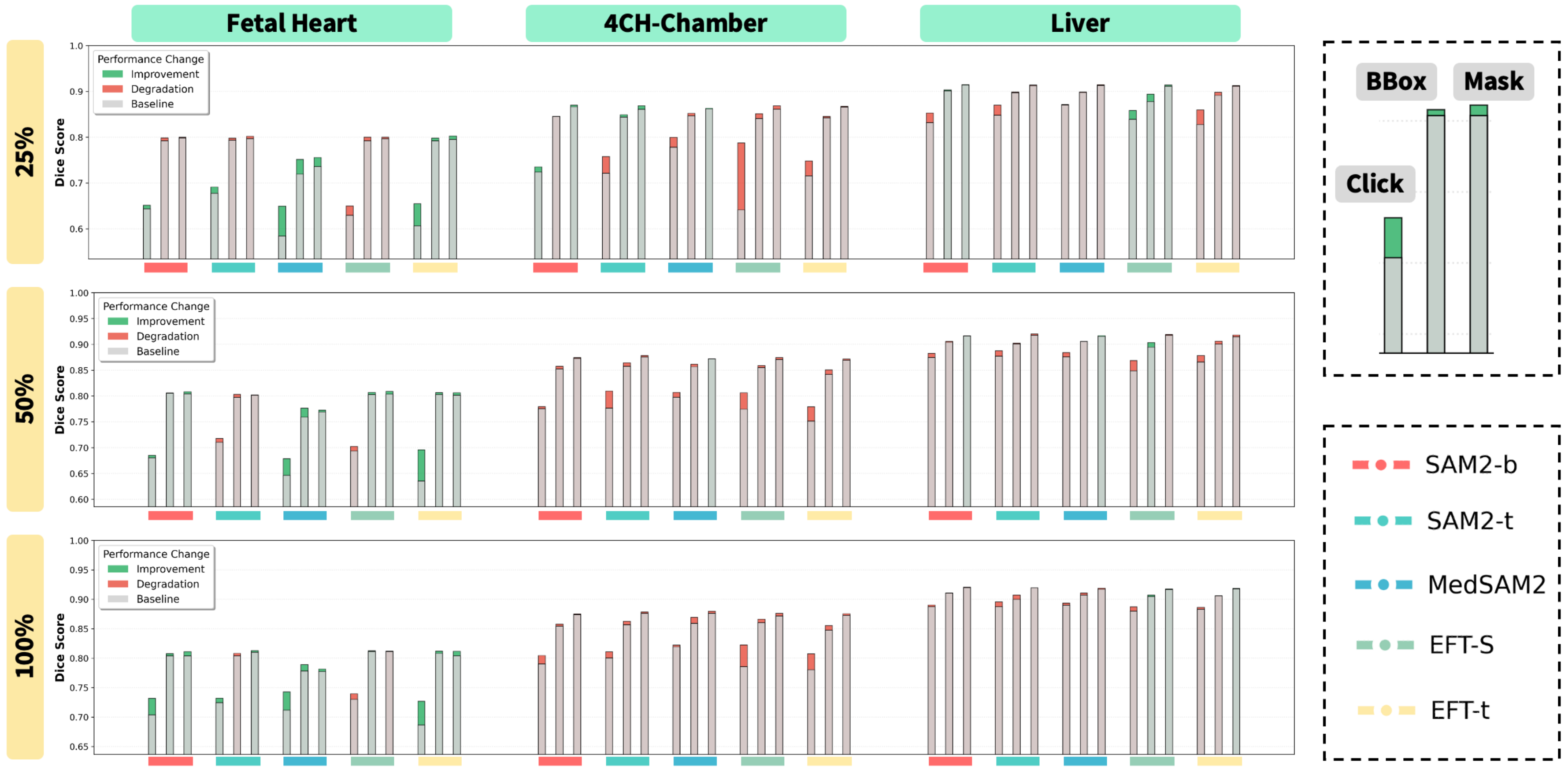}
    \caption{Joint training (JT) results on the \textit{core target datasets} under different training scales.}
    \label{fig:joint_train}
\end{figure*}


\subsection{How does MT/MT-FT help with task-specific segmentation?}
\label{subsec:midtrainzeroshot}

\subsubsection{Implementation Details}

As described in \textbf{Section~\ref{subsec:midtraining}}, we evaluate the performance of MT and MT-FT on three \textit{core target datasets} individually. During MT, each model is pretrained on four \textit{auxiliary datasets} and the two remaining \textit{core target datasets}, excluding the target dataset used for testing to avoid data leakage. Both long and real-world short training videos (ST) are included in this stage. The pretrained MT models are then evaluated on the held-out \textit{core target dataset} in a zero-shot setting. For reference, we also measure the zero-shot performance of each model before MT. In the MT-FT phase, the pretrained MT models are further FT on their respective \textit{core target datasets}. We examine three training scales:  
(i) 25\% (\text{T3}) of the long training videos, representing a limited-data setting;  
(ii) 100\% (\text{T1}) of the long training videos; and  
(iii) the combination of all long and short training videos (\text{T1+ST}), representing a data-sufficient scenario.  For each scale, FT-only results without MT are used as baselines for comparison.

\subsubsection{Results}

Figure~\ref{fig:midtrain} summarizes the evaluation results. The first row presents the zero-shot performance of models with MT compared to the original foundation models. The second to fourth rows show the MT-FT performance compared to the FT-only baseline across three training scales. Each column corresponds to one \textit{core target dataset}. The gray bars indicate the baseline results, the green bars represent improvements obtained through MT or MT-FT, and the red bars denote performance drops relative to the baseline.

\noindent\textbf{MT vs. zero-shot:} In the zero-shot segmentation results (Figure~\ref{fig:midtrain}, first row), MT provides a clear improvement under the Click-prompting setting. However, for Box-prompting, performance degradation dominates in the Fetal Heart and 4-Chamber datasets. For Mask-prompting, the results are inconsistent, showing either minor degradation or no effect across models and datasets. Since Click-prompting consistently yields the lowest baseline performance, the observed improvement in this setting offers limited practical value.

\noindent\textbf{MT-FT vs. FT:}  
When comparing MT-FT with FT-only across different training scales, both improvements and degradations appear inconsistently across datasets, models, and prompting strategies. We also observe that as the fine-tuning scale increases, the influence of MT gradually diminishes. This trend is evident from the second to the fourth rows in Figure~\ref{fig:midtrain}.

\noindent\textbf{Practical Summary:}  
Considering the limited effectiveness and high computational cost of MT, our results indicate that both MT and MT-FT can provide little practical benefit when the goal is to achieve strong performance on a specific downstream task. In such cases, direct FT is more efficient and equally effective. This finding is consistent with our zero-shot results, where the \textbf{Medical Groups} underperform the \textbf{Non-Medical Groups}, and also aligns with the conclusions of \cite{gu2024build}, who reported similar observations in their study on task-expansive fine-tuning of SAM.


\subsection{How does joint training help with task-specific segmentation?}
\label{subsec:jointtrain}

\subsubsection{Implementation Details}


As discussed in \textbf{Section.\ref{subsec:jointtraining}}, we systematically evaluate the effectiveness of JT for ultrasound video segmentation. We benchmark JT against FT on different training scales (25$\%$, 50$\%$, 100$\%$). For each training scale, the JT is performed on the combination of long training videos from all 3 \textit{core target datasets}. 

\subsubsection{Results}

Figure~\ref{fig:joint_train} illustrates the results of JT. From the top to the bottom row, the training scale is increasing from 25$\%$ to 100$\%$, simulate the training regimes from data-limited situation to data-sufficient situation. Gray bars denote baseline performance of FT while improvements and degradations of using JT are highlighted in green and red, respectively. 

\noindent\textbf{Effect of Training Scale:}  
Across different training scales (25$\%$ to 100$\%$), the effect of joint training (JT) does not follow a clear monotonic trend. Both improvements and degradations appear inconsistently across datasets, models, and prompting strategies. For a fixed dataset–model pair, JT tends to have a stronger impact under Click prompting than under BBox or Mask prompting. However, performance patterns remain largely consistent across scales, showing only minor variation. For instance, EFT-t consistently benefits from JT on the Fetal Heart dataset at all scales, while it shows consistent degradation on 4-Chamber, regardless of data size. These observations suggest that the effectiveness of JT depends more on dataset- and task-specific factors than on training data scale.

\noindent\textbf{Model-Level Variability:}  
Models respond differently under similar JT conditions. For example, in the Fetal Heart dataset, both MedSAM2 and EFT-t show clear improvements, while other models exhibit mixed results with occasional gains or losses. Although model behaviors vary within the same dataset, these differences are relatively small. More importantly, no single model achieves consistent improvement across datasets; a model that benefits from JT on one dataset may lose this advantage on another.

\noindent\textbf{Dataset-Level Variability:}  
The dataset is the most consistent factor influencing JT performance. Overall, JT leads to greater improvements on the Fetal Heart dataset compared with the other two. Dataset-specific patterns are also evident for individual models. For example, EFT-s experiences a clear performance drop after JT on 4-Chamber but shows noticeable gains on Liver. These results suggest that dataset characteristics are the primary factors shaping JT effectiveness, while model differences have a smaller influence.

\noindent \textbf{JT vs. FT:} Compared with dataset-specific fine-tuning, JT occasionally brings improvements but more often results in moderate performance declines. Most degradations are minor, except for EFT-s under Click prompting in the 4-Chamber dataset, where performance drops markedly.

\noindent\textbf{Practical Summary:}  
From a practical perspective, FT remains the preferred approach for achieving optimal performance on a specific task. However, JT can be a practical alternative when multiple datasets are available. Although JT may not always match the task-specific accuracy of FT, it offers clear advantages in computational efficiency, storage, and inference unification. These benefits make JT a viable choice for large-scale multi-dataset applications.

\begin{figure*}[h]
    \includegraphics[width=1\textwidth]{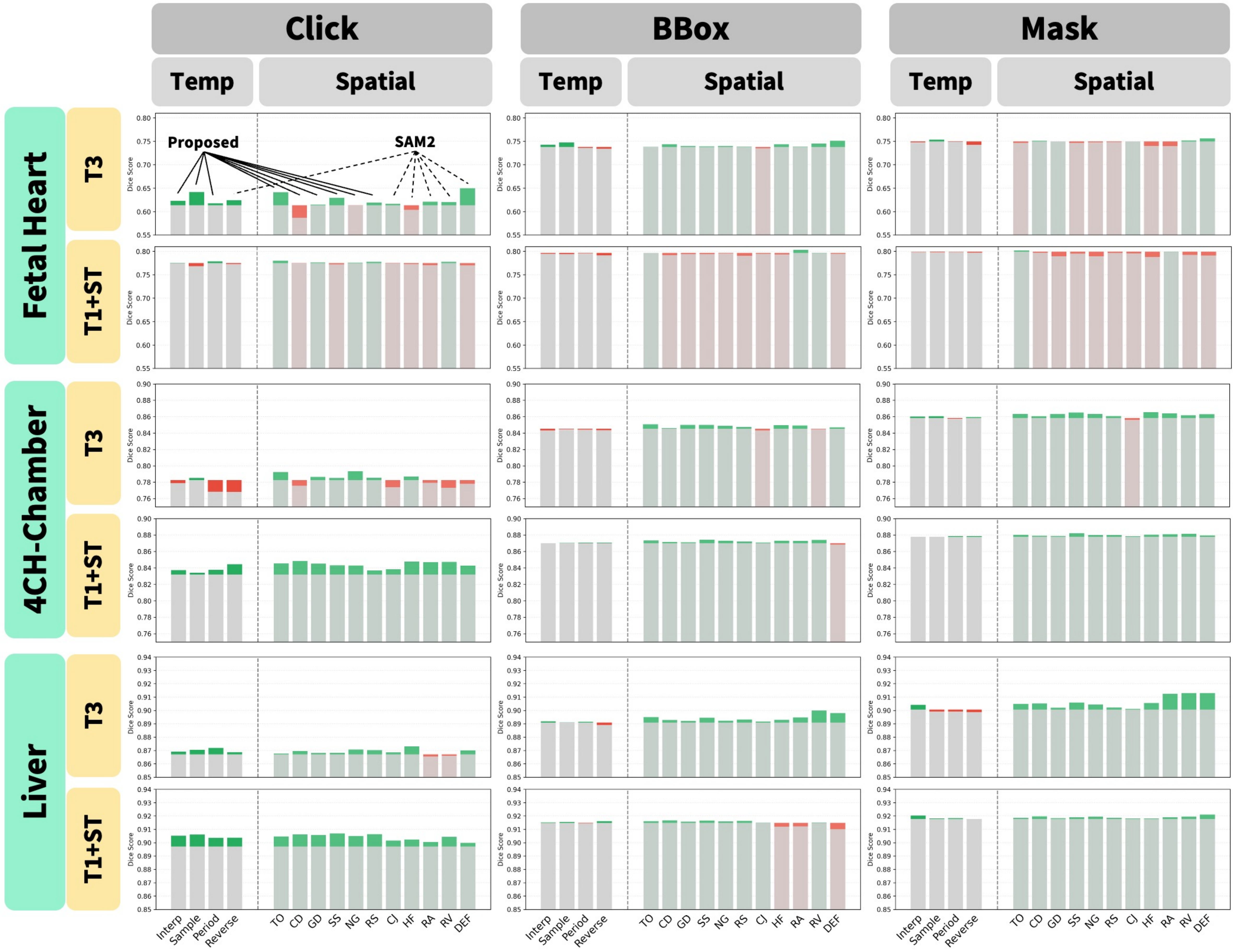}
\caption{
Joint training (JT) results with temporal and spatial augmentations on the \textit{core target datasets} under different training scales. 
Augmentation abbreviations: \textit{TO}: Text Overlay; \textit{CD}: Contrast Degradation; \textit{GD}: Gaussian Decay; 
\textit{SS}: Shadow Simulation; \textit{NG}: Noise with Gaussian Degradation; \textit{RS}: Random Structure Injection; 
\textit{CJ}: Color Jitter; \textit{HF}: Horizontal Flip; \textit{RA}: Random Affine; \textit{RV}: Random Video Mosaic; \textit{DEF}: Default SAM2 augmentation.
}

    \label{fig:aug_results}
\end{figure*}

\subsection{The influence of spatial and temporal augmentations}
\label{subsec:s_aug_train}

\subsubsection{Implementation Details}

We evaluate the effects of different spatial and temporal augmentation methods described in \textbf{Section~\ref{subsec:aug}}. 
For computational efficiency, each augmentation type is assessed through JT on the \text{core target datasets} using three ratios or probabilities (0.2, 0.5, and 0.8). We retain the original hyperparameters of the \textit{SAM2 default combined augmentation (DEF)}.
For each method, the result with the best-performing ratio is reported.  We conduct experiments under two settings:  
(i) 25\% of long training videos (\text{T3}), representing limited data; and  
(ii) 100\% of both long and short training videos (\text{T1+ST}), representing sufficient data.
MedSAM2 is used as both the backbone and initialization model, as it combines the SAM2-t architecture with medical-domain pretraining and shows strong joint training performance.

\subsubsection{Results – General Analysis}
Figure~\ref{fig:aug_results} summarizes JT performance under different augmentation strategies. Solid lines indicate the proposed ultrasound-specific augmentations, whereas dashed lines represent SAM2’s built-in augmentations. For each method, green and red bars denote performance improvements and degradations relative to the no-augmentation baseline, respectively.

\paragraph{\textbf{Prompting Mode Differences:}}  
Consistent with previous experiments, the effect of augmentation is most pronounced under \text{Click} prompts and relatively weak under \text{BBox} and \text{Mask}. 
This can be explained by the sparse and local nature of Click prompts, which forces the model to rely more on temporal and spatial coherence for consistent segmentation across frames. 
In contrast, Box and Mask prompts provide stronger spatial priors, reducing the benefit gained from augmentation-induced consistency.

\paragraph{\textbf{Context Sensitivity:}}  
The impact of augmentations is highly dataset-dependent. 
For 4-Chamber and Liver datasets, most augmentations yield performance gains, whereas for Fetal Heart, many methods have limited or even adverse effects. 
The relationship between augmentation efficacy and training scale is also non-monotonic. 
For example, in Fetal Heart, some augmentations cause performance to deteriorate under \text{T1+ST} but show steady improvement under the smaller \text{T3} setting.
Conversely, in 4-Chamber, nearly all augmentations lead to noticeable improvements under \text{T1+ST} with Click-prompting, yet several temporal and SAM2 built-in sptial augmentations exhibit negative effects at\text{T3}. 
These findings indicate that the effectiveness of augmentation is shaped by both dataset characteristics and training scale, though no simple or universal trend can be inferred.


\subsubsection{Results - Temporal Augmentations}
Temporal augmentations have the strongest effect under \text{Click} prompting and moderate influence under \text{BBox} and \text{Mask}. 
With Click prompting, all four temporal strategies improve performance on Fetal Heart (\text{T3}), 4-Chamber (\text{T1+ST}), and Liver (\text{T1+ST, T3}). 
This suggests that interpolation, periodic repetition, and downsampling enhance robustness to motion variability by increasing temporal diversity. 
In contrast, the \textit{Reverse} augmentation consistently lowers performance on Fetal Heart (\text{T1+ST}) and 4-Chamber (\text{T3}), likely because reversing frame order disrupts physiologically meaningful motion. 
For relatively static organs such as the Liver, the impact of reversal is minimal. 
Overall, temporal augmentations have a moderate yet context-dependent effect, largely shaped by the prompting mode and the temporal dynamics of the target anatomy.

\subsubsection{Results – SAM2 Buit-in Spatial Augmentations}

Compared to temporal augmentations, spatial transformations yield more consistent performance gains across datasets, with the only notable exception observed in Fetal Heart (\text{T1+ST}). 
The limited benefit in this case may stem from the task’s inherently complex anatomical boundaries and rapid motion, where spatial perturbations could amplify segmentation difficulty rather than improve robustness.

\paragraph{\textbf{Photometric Augmentations:}}  

Generally, \textit{Color Jitter (CJ)} exhibits a slight but consistent positive effect on ultrasound segmentation performance for the 4CH-Chamber and Liver datasets, while showing a moderate negative effect on the Fetal Heart dataset. The SAM2 built-in \textit{CJ} function includes four operations—contrast, brightness, saturation, and hue adjustments. However, since ultrasound images are inherently grayscale, \textit{CJ} does not affect saturation or hue, influencing only contrast and brightness. Interestingly, we observe that \textit{CJ} follows a similar trend to our proposed ultrasound-specific \textit{Contrast Degradation (CD)} augmentation, which is reasonable given that both focus on modifying image contrast. Overall, \textit{CJ} performs slightly worse than \textit{CD} across most datasets.

\paragraph{\textbf{Geometric Transformations:}}  
\textit{Random Affine (RA)} generally improves performance under \text{BBox}- and \text{Mask}-prompting but less so under \text{Click}-prompting. 
This can be attributed to the stronger spatial priors of BBox and Mask prompts, which allow the model to tolerate affine perturbations, while Click-prompting is more sensitive to geometric misalignment between the prompt and the target structure. 
\textit{Horizontal Flip (HF)} yields relatively consistent effects across prompting modes, as it preserves anatomical structure while altering only orientation. 
However, in clinical applications, HF should be used with caution since left–right inversion may conflict with medically meaningful laterality and reduce interpretability.

\paragraph{\textbf{Random Video Mosaic:}}  
The \textit{Random Video Mosaic (RV)} introduces spatial diversity by compositing multiple downscaled frames into a single mosaic, with the target mask retained only in one randomly selected grid. This strategy effectively enriches spatial variation and improves model robustness to object position, scale, and contextual complexity, which is particularly beneficial in video segmentation tasks. However, since other grids may contain unlabeled instances of the same object, it can inadvertently introduce false negatives.In practice, using a moderate application probability and consistent mask placement across frames balances these trade-offs, allowing the model to gain positional robustness while mitigating the risk of label noise. In our experiments, the mild improvements observed suggest that prompt conditioning and low occurrence frequency mitigate potential label noise, making RV a safe yet limited augmentation that provides moderate regularization without substantial gains.

\paragraph{\textbf{Combined Augmentation Effects:}}  
Comparing the \textit{SAM2 default combined augmentation (DEF)} with its individual components reveals non-linear interactions among augmentation methods. 
For example, in Fetal Heart (\text{T3}), the combined setting yields clear improvement despite several individual augmentations showing neutral or negative effects. 
Conversely, in 4-Chamber (\text{T1+ST}), the same combination results in degradation even though individual augmentations perform well. 
These observations indicate that augmentation strategies can interact in either complementary or conflicting ways, underscoring the importance of evaluating them not only in isolation but also in combination.


\subsubsection{Results - Proposed US-Specific Spatial Augmentations:}  
The proposed spatial augmentations lead to more consistent performance gains than the built-in SAM2 augmentations, particularly in the 4-Chamber and Liver datasets, where all methods except \textit{Contrast Degradation (CD)} show improvements. 
In Fetal Heart, the proposed augmentations outperform SAM2 augmentations under \text{T3}, while results under \text{T1+ST} remain comparable.

\paragraph{\textbf{Structure Perturbations:}}Augmentations introducing structural perturbations, such as \textit{Text Overlay (TO)}, yield clear benefits even in datasets without textual content, suggesting that localized structural disturbances improve robustness to occlusions and appearance variations. 
Similarly, \textit{Shadow Simulation (SS)} and \textit{Random Structure Injection (RS)} consistently enhance performance, likely by mimicking realistic acoustic shadows and tissue heterogeneity that promote cross-condition generalization.

\paragraph{\textbf{Intensity Perturbations:}}  \textit{Gaussian Decay (GD)} and \textit{Noise with Gaussian Decay (NG)} exhibit nearly identical trends across all settings, as both simulate depth-dependent attenuation—one of ultrasound’s core physical characteristics. 
While NG adds Gaussian noise atop the decay field, its contribution remains secondary to the attenuation effect, indicating that \textit{GD} alone captures most of the practical benefit. 
These results suggest that augmentations based on similar physical principles tend to saturate in effect, and future designs should emphasize introducing distinct imaging factors rather than redundant variations.

\textit{Contrast Degradation (CD)} produces limited impact, with mostly weak positive or neutral effects. 
As the transformation primarily modifies global intensity mapping rather than structural boundaries, it offers minimal benefit for segmentation tasks that rely on spatial consistency. 
Nonetheless, mild gains are observed in cases involving varying probe contrast, supporting \textit{CD}’s role as a low-risk but low-impact augmentation.


\subsubsection{Practical Summary}

Overall, our experiments show that data augmentation exerts a generally positive but modest influence on fine-tuning ultrasound foundation models, with a smaller effect compared to factors such as training scale. 
Temporal augmentations yield weak and task-dependent improvements, whereas spatial augmentations guided by clinically realistic priors achieve more consistent gains by directly emulating acquisition physics and anatomical variability. 
These findings suggest a pragmatic strategy for future research: augmentation should be treated as a secondary factor during early-stage model exploration to reduce computational cost, while later optimization can prioritize spatial techniques grounded in imaging physics and anatomical plausibility. 

\section{Conclusion and Discussion}
\label{sec:conclusion}
This study systematically investigates SAM2 adaptation for ultrasound video segmentation, revealing that data characteristics and adaptation strategies dominate over architectural complexity in determining performance. Scaling data proves more effective than scaling parameters, while short videos enhance learning mainly in data-scarce settings. Direct fine-tuning remains the most efficient strategy for task-specific optimization, whereas joint training offers a practical trade-off when multiple datasets are available.  Spatial augmentations informed by imaging physics yield more consistent gains than generic temporal transformations, underscoring the value of physically grounded design.

It is important to note that our analysis focuses exclusively on ultrasound video data, chosen to emphasize temporal characteristics rather than 3D volumetric imaging. Moreover, due to institutional data governance policies, only internal datasets were used for experimentation, which may constrain the diversity of imaging conditions and patient populations. Despite these limitations, the consistent trends observed across tasks suggest that the insights are broadly applicable to medical video segmentation. Future research should expand toward multi-modal and volumetric data, integrate self-supervised and cross-domain adaptation schemes, and improve interpretability and uncertainty estimation to achieve scalable and clinically trustworthy medical foundation models.

\acks{


We acknowledge the support of Dr.Gopal Avinash and Dr.Ipek Oguz in finishing this work. We also acknowledge our colleagues, Pavan Annangi and Harsh Suthar for making the data available to us, and Dwijay D Shanbhag, Dr.Satrajit Chakrabarty, Dr.Keerthi Sravan Ravi, Dr.Fei Tan, Ashok Vardhan Addala, Md Mostafijur Rahman, Tonmoy Hossain, and Zelong Liu for their insights. We are also grateful to Haoyu Dong for the helpful conversations.

}

%
\ethics{The work follows appropriate ethical standards in conducting research and writing the manuscript, following all applicable laws and regulations regarding treatment of animals or human subjects.}

\coi{All authors were employed by GE Healthcare during the research and
manuscript preparation. However, the research was conducted independently, and the authors declare that the results and conclusions are presented
objectively, without external influence from their employer.}

\data{The data used in this study are internal and cannot be publicly released due to privacy considerations and company regulations. All methods employed in this work are based on publicly available open-source implementations, and we can provide the complete list of hyperparameters upon request. The proposed augmentation methods are currently internal techniques; however, their implementation principles and detailed documentation are available in the \textbf{Appendix} for reproducibility purposes.}

\bibliography{sample}


\clearpage

\end{document}